\documentclass{article}
\usepackage{iclr2027_conference,times}

\usepackage{hyperref}
\usepackage{url}
\usepackage{booktabs}
\usepackage{graphicx}
\usepackage{subcaption}
\usepackage{amsmath,amssymb}
\usepackage{multirow}
\usepackage{algorithm}
\usepackage{algpseudocode}
\usepackage{microtype}
\usepackage{enumitem}

\newcommand{\physsae}{\textsc{PhysSAE}}
\newcommand{\ESF}{\mathrm{ESF}_{80}}
\newcommand{\pnf}{p_{95}}

\title{\physsae: Mechanistic Interpretability of PINNs with Sparse Autoencoders}

\author{%
    Nandita N. Patil$^{*}$\thanks{\href{mailto:nanditanpatil114@gmail.com}{\texttt{nanditanpatil114@gmail.com}}},
    Eshwar R A$^{*}$\thanks{\href{mailto:eshwarra5@gmail.com}{\texttt{eshwarra5@gmail.com}}},
    and Gajanan V Honnavar$^{\dagger}$\thanks{\href{mailto:gajanan.v@qnulabs.com}{\texttt{gajanan.v@qnulabs.com}}}
    \\[2mm]
    $^{*}$QuaNad Research Laboratory, PES University, Bengaluru, India\\
    $^{\dagger}$Independent Researcher, QNu Labs Pvt Ltd, Bengaluru, India. Former Professor at \\PES University (EC Campus), Bengaluru
}

\iclrfinalcopy
\begin{document}
\begingroup
\renewcommand{\fnsymbol}[1]{(\alph{#1})}
\maketitle
\endgroup
\lhead{Preprint---\physsae: Mechanistic Interpretability of PINNs with Sparse Autoencoders}

\begin{abstract}
Physics-informed neural networks (PINNs) solve PDEs by minimising the equation residual, yet what their hidden
layers represent is largely unknown. We present \physsae, which trains sparse autoencoders (SAEs) on PINN
activations, aligns the learned atoms with physical fields computed from independent reference solutions, and
tests them by editing the frozen network's hidden state. Across 12 PDEs and 36 PINNs, atoms form readable
shock, interface, soliton and front features (best $|r|$ up to 0.98), and removing an atom changes the solution
mostly where its concept lives. Comparing every statistic with untrained networks and with PINNs trained on
other equations shows that trained PINNs encode the individual terms of their own PDE operator. A pre-registered
test on three unseen equations confirmed all five predictions (Burgers diffusion term: probe $R^2=0.99$ against
0.15 for the null networks). PINNs that fail to train lose this structure, which makes \physsae{} a diagnostic of
what a PINN has learned.
\end{abstract}

\section{Introduction}
\label{sec:intro}

Physics-informed neural networks~\citep{raissi2019physics} encode the residual of a PDE directly into the training
loss via automatic differentiation, enabling mesh-free solution of forward and inverse problems. They are now used
for heat transfer~\citep{cai2021physics}, solid mechanics~\citep{haghighat2021physics} and biomedical
modelling~\citep{sahli2020physics}, but they remain black boxes at the feature level. Their failure modes
(spectral bias~\citep{wang2022ntk}, gradient pathologies~\citep{wang2021gradflow}, violations of temporal
causality~\citep{wang2024causal}) are well characterised at the level of the loss and its gradients. Much less is
known about what a trained PINN represents in its hidden layers. A direct inspection of PINN weights
by~\citet{tucny2025insight} found weight distributions close to Gaussian, with no coarse trace of the physics,
which suggests looking at features rather than weights.

Mechanistic interpretability has developed tools for exactly this: linear probes~\citep{alain2017probing}, concept
activation vectors~\citep{kim2018tcav}, and overcomplete sparse autoencoders
(SAEs)~\citep{bricken2023monosemanticity,cunningham2023sae,templeton2024scaling}. An SAE rewrites each hidden
activation vector as a sparse, non-negative combination of a larger set of learned directions, called
\emph{atoms}, in the hope that each atom corresponds to one human-readable feature. SAEs have recently been
applied to graph CFD surrogates~\citep{hu2025cfdsae} and neural operators~\citep{tolooshams2025saeno}, and concept
steering has been demonstrated on a physics foundation model~\citep{fear2025physteer}.

\physsae{} brings these tools to residual-loss PINNs, and the answer it gives is encouraging. SAE atoms trained on
PINN activations form features a physicist would recognise: the Burgers shock, the Allen--Cahn interface, the
moving Schr\"odinger soliton and its real and imaginary parts, the Fisher--KPP front. More importantly, trained
PINNs encode the individual terms of their own PDE operator. In the Burgers PINN, a linear read-out of the last
hidden layer recovers the diffusion term $\nu u_{xx}$ with $R^2=0.66$, while 95\% of networks that were never trained
on Burgers stay below 0.08. We stated this as a prediction before training PINNs on three new equations, and it
held for all of them. Removing a concept-aligned atom from the hidden state changes the PINN's solution mostly
inside the region where that concept is large, and PINNs that fail to solve their equation do not encode its
derivative structure.

Three design choices make these statements possible in the PINN setting. First, concepts are computed from
independent reference solutions rather than from the PINN's own output, so alignment cannot be produced by the
readout. Second, interventions act on the original hidden state of the frozen PINN rather than on an SAE
reconstruction. Third, every statistic is compared with the same statistic computed on \emph{null networks}
(untrained networks of the same architecture, and PINNs trained on other PDEs). Smooth functions of $(x,t)$ are
easy for any network to represent, so this comparison is what separates structure produced by training from
structure that any network of the same shape would show~\citep{hewitt2019control,heap2026random}.

\paragraph{Related work.}
\textbf{PINN architectures and failure modes.}
Self-adaptive weights~\citep{mcclenny2020sapinn} and Fourier-feature embeddings~\citep{wang2021fourier} address
spectral bias; domain decomposition~\citep{jagtap2020xpinn} enables parallel training; causal
reweighting~\citep{wang2024causal} enforces temporal ordering; the failure of PINNs on stiff reaction and
high-speed convection problems was characterised by~\citet{krishnapriyan2021failures}; and KAN-based
PINNs~\citep{wang2024kinn} offer symbolic readability through spline activations. None of these study internal
representations at the feature level.
\textbf{Interpretability in scientific ML.}
Attribution methods such as SHAP~\citep{lundberg2017shap} treat a PINN as a black-box regressor and do not engage
with its spatio-temporal structure. SAEs on graph CFD surrogates~\citep{hu2025cfdsae} and neural
operators~\citep{tolooshams2025saeno}, and concept steering in a physics foundation model~\citep{fear2025physteer},
are closest to our work, but they study data-driven models rather than residual-loss PINNs.
\textbf{Sparse autoencoders and their evaluation.}
We follow the standard SAE design (ReLU codes, L1 sparsity, unit-norm decoder
columns)~\citep{bricken2023monosemanticity,cunningham2023sae,templeton2024scaling,elhage2021framework} and compare it
with TopK SAEs~\citep{gao2024topk}. Probing studies use control tasks to check that a probe reads out the model
rather than learning the task itself~\citep{hewitt2019control,belinkov2022probing}, and SAEs trained on randomly
initialised transformers can yield latents that score as well on standard interpretability metrics as those of trained
models~\citep{heap2026random}. Our null networks play the same role for PINNs.

\paragraph{Contributions.}
\begin{enumerate}[leftmargin=*,itemsep=2pt]
\item \textbf{An SAE framework for PINN latent spaces} with an independent-concept protocol that avoids circular
validation, applied to 12 PDEs ranging from smooth diffusion to shocks, interfaces, fronts and solitons.
\item \textbf{Direct hidden-state intervention}, $h_\ell\to h_\ell-\alpha\,z_k\,(\sigma_\ell\odot d_k)$ at any hidden
layer, with effects measured on the PINN solution and on its PDE residual and compared with PCA, ICA and matched
control atoms.
\item \textbf{Null-calibrated evaluation}: untrained networks and PINNs trained on other PDEs give a reference
distribution for every alignment, probing, two-atom and intervention statistic.
\item \textbf{Operator-term encoding}: trained PINNs represent the terms of their own operator far beyond any null
network, confirmed by a pre-registered test on three unseen PDEs.
\item \textbf{A practical toolkit} for reading individual features: distinct-atom assignment, selectivity
scores and spatial profiles that tell apart atoms whose activation maps look alike, plus a diagnostic of what a
failed PINN has and has not learned.
\end{enumerate}

\paragraph{Paper overview.}
Section~\ref{sec:method} describes the method, Section~\ref{sec:setup} the PDE benchmark,
Section~\ref{sec:results} the results, and Section~\ref{sec:discussion} their implications and limitations.

\section{Method}
\label{sec:method}

\physsae{} is a post-hoc framework that operates on a \emph{trained, frozen} PINN; no retraining or architectural
change is needed. It (1)~records the PINN's hidden activations on a dense space-time grid, (2)~trains a sparse
autoencoder on them, (3)~aligns the resulting atoms with physical concept fields, and (4)~tests atoms by editing
the original hidden state and measuring how the PINN's output and PDE residual change. Every statistic is also
computed on a set of null networks (\S\ref{sec:nulls}). We describe each step below and define every quantity we
report, so that the results can be read without consulting the cited works.

\subsection{Problem setting and notation}
\label{sec:notation}

Let $\Omega\subset\mathbb{R}$ be a spatial interval and $[0,T]$ a time interval. We consider PDEs
\begin{equation}
  \mathcal{F}[u](x,t) = 0 \qquad \text{on } \Omega \times (0,T],
  \label{eq:pde}
\end{equation}
with initial condition $u(x,0)=u_0(x)$ and boundary condition $\mathcal{B}[u](x,t)=0$ on $\partial\Omega\times[0,T]$.
We write $u_\theta:\Omega\times[0,T]\to\mathbb{R}^{d_{\mathrm{out}}}$ for the PINN, with $d_{\mathrm{out}}=1$ for real
PDEs and $d_{\mathrm{out}}=2$ for the nonlinear Schr\"odinger equation, whose complex field $\psi=\psi_r+i\psi_i$ is
split into real and imaginary parts. (We reserve $h$ for hidden states throughout.)

The PINN is a fully connected network with $L$ hidden layers of width $d_h$ and a linear readout:
\begin{equation}
  u_\theta(x,t) = W_{\mathrm{out}}\, h_L(x,t) + b_{\mathrm{out}},
  \qquad h_\ell = \tanh(W_\ell h_{\ell-1} + b_\ell),
  \quad h_0 = [x,t]^\top.
  \label{eq:pinn_arch}
\end{equation}
We call $h_\ell(x,t)\in\mathbb{R}^{d_h}$ the \emph{hidden state} of layer $\ell$ at the point $(x,t)$. Most analyses
use the last hidden layer $h\equiv h_L$, the richest representation before the linear readout. Because the output
is linear in $h_L$, any change to $h_L$ reaches the output through $W_{\mathrm{out}}$ alone; changes made at earlier
layers pass through the remaining nonlinear layers first. We use both properties when choosing where to intervene
(\S\ref{sec:intervention}).

\subsection{PINN training}
\label{sec:pinn_training}

\paragraph{Architecture.}
We use $L=5$ hidden layers of width $d_h=128$, $\tanh$ activations and Xavier-normal
initialisation~\citep{glorot2010}. The size is moderate on purpose: large enough that a layer can hold more
features than it has units, small enough that the 36 PINNs in this study (12 PDEs $\times$ 3 seeds) train on a
single GPU in a few hours.

\paragraph{Training objective.}
The PINN minimises
\begin{equation}
  \mathcal{L}(\theta)
  = w_{\mathrm{BC}}\,\mathcal{L}_{\mathrm{BC}}
  + w_{\mathrm{IC}}\,\mathcal{L}_{\mathrm{IC}}
  + w_F\,\mathcal{L}_F,
  \label{eq:pinn_loss}
\end{equation}
where each term is a mean over sampled points:
\begin{align}
  \mathcal{L}_F &= \frac{1}{N_F}\sum_{i=1}^{N_F}
    \|\mathcal{F}[u_\theta](x_i^F, t_i^F)\|_2^2,
  \label{eq:res_loss}\\
  \mathcal{L}_{\mathrm{IC}} &= \frac{1}{N_{\mathrm{IC}}}\sum_{i=1}^{N_{\mathrm{IC}}}
    \|u_\theta(x_i,0) - u_0(x_i)\|_2^2,\\
  \mathcal{L}_{\mathrm{BC}} &= \frac{1}{N_{\mathrm{BC}}}\sum_{i=1}^{N_{\mathrm{BC}}}
    \|\mathcal{B}[u_\theta](x_i^B, t_i^B)\|_2^2.
\end{align}
The residual $\mathcal{F}[u_\theta]$ is evaluated exactly by automatic differentiation through the network. We use
$w_{\mathrm{BC}} = w_{\mathrm{IC}} = 100$ and $w_F = 1$, a common choice that makes the network honour the initial and
boundary data even when the residual is stiff. Each step samples $N_F = 3000$ residual points, $N_{\mathrm{IC}} = 150$
initial points and $N_{\mathrm{BC}} = 150$ boundary points uniformly at random.

\paragraph{Optimisation schedule.}
Training has two stages. Adam~\citep{kingma2015adam} with learning rate $10^{-3}$ runs for 8000 iterations with
gradient clipping at norm 1.0, which prevents instabilities on the stiff Allen--Cahn and high-speed convection
problems. L-BFGS~\citep{liu1989lbfgs} with strong-Wolfe line search then runs for up to 300 iterations; this
quasi-Newton method uses curvature information to sharpen the solution once Adam has found a good basin. If
L-BFGS increases the loss, the Adam solution is kept.

\paragraph{Boundary conditions.}
For Dirichlet conditions $u(x_{\mathrm{bnd}},t)=g(x_{\mathrm{bnd}},t)$ we sample points on $\partial\Omega\times[0,T]$
and penalise the squared mismatch. For periodic conditions we enforce periodicity of both the value and its first
derivative:
\begin{equation}
  \mathcal{L}_{\mathrm{BC}}^{\mathrm{per}}
  = \frac{1}{N}\sum_i
    \bigl(\|u_\theta(a,t_i)-u_\theta(b,t_i)\|^2
    + \|\partial_x u_\theta(a,t_i) - \partial_x u_\theta(b,t_i)\|^2\bigr),
  \qquad \Omega=[a,b].
\end{equation}

\subsection{Activation extraction}
\label{sec:activation_extraction}

After training, the PINN is frozen and evaluated on a uniform grid
$\mathcal{G} = \{(x_i, t_j)\}$ with $n_x=200$ and $n_t=100$ points ($N=20{,}000$). At every grid point we record all
five hidden states, giving one activation matrix per layer,
\begin{equation}
  H_\ell = \bigl[h_\ell(x_1,t_1),\,\ldots,\,h_\ell(x_{n_x},t_{n_t})\bigr]^\top \in \mathbb{R}^{N \times d_h}.
  \label{eq:H}
\end{equation}
Each row describes the network's internal state at one space-time point.

\paragraph{Normalisation.}
Each hidden unit is z-scored over the grid,
\begin{equation}
  \tilde{h}_k = \frac{h_k - \mu_k}{\sigma_k + \varepsilon},
  \qquad
  \mu_k = \frac{1}{N}\sum_{(x,t)\in\mathcal{G}} h_k(x,t),
  \quad
  \sigma_k^2 = \frac{1}{N}\sum_{(x,t)\in\mathcal{G}} (h_k(x,t) - \mu_k)^2,
  \label{eq:zscore}
\end{equation}
with $\varepsilon=10^{-6}$. This puts all units on the same scale, so the sparsity penalty of the SAE treats them
alike, and it makes the unit-norm constraint on decoder directions meaningful. We keep $\sigma=(\sigma_1,\dots,\sigma_{d_h})$
to map SAE directions back to raw hidden coordinates when we intervene (\S\ref{sec:intervention}).

\paragraph{Train/evaluation split.}
Grid points are split at random into 80\% (16{,}000 points) for fitting SAEs and probes and 20\% (4{,}000 points)
for evaluation. Every alignment, probe and intervention statistic in this paper is computed on the held-out 20\%.

\subsection{Independent reference solutions and concept fields}
\label{sec:concepts}

A central principle of \physsae{} is that concept fields come from sources \emph{independent} of the PINN. The PINN
output is a linear function of its last hidden state, so any field derived from $u_\theta$ would correlate with
$h_L$ whether or not the PINN had learned the physics. We therefore compute concepts from:
\begin{itemize}[leftmargin=*,itemsep=1pt]
  \item \textbf{Exact solutions} where they exist: the heat equation with Gaussian initial data,
    $u=(1+20t)^{-1/2}\exp(-5x^2/(1+20t))$; linear convection, $u=u_0(x-\beta t)$ by the method of characteristics; the
    Schr\"odinger soliton~\citep{zakharov1972}; the KdV soliton; the Fisher--KPP travelling
    front~\citep{ablowitz1979fisher}; and advection--diffusion with an error-function front.
  \item \textbf{Fourier pseudo-spectral solutions} for the Burgers and Allen--Cahn problems: an integrating-factor
    fourth-order Runge--Kutta scheme~\citep{lawson1967} with $2/3$ de-aliasing and 2048--8192
    modes~\citep{trefethen2000spectral}, interpolated onto $\mathcal{G}$. Each solution agrees with one at half the
    resolution to within $3\times10^{-5}$ in relative $L^2$. References are computed once and shared by all seeds.
\end{itemize}

\paragraph{Concept fields.}
From the reference $u^*$ we compute a small panel of scalar fields on $\mathcal{G}$, using central differences for
derivatives (Table~\ref{tab:concepts}). The panel covers four kinds of quantity: the solution itself (signed and
unsigned), its derivatives (where it is steep or curved), PDE-specific quantities (interface indicator,
characteristic coordinate), and the \emph{operator terms}, which are the individual terms of $\mathcal{F}[u^*]$
evaluated on the reference. For Burgers these are the advection term $u^*u^*_x$ and the diffusion term $\nu u^*_{xx}$.
They are the quantities a PINN has to compute and balance to make its residual vanish, which makes them the most
direct test of whether the network represents its equation. Fields that are identical up to scale were merged
(for the heat equation $u_t=u_{xx}$; for a rigidly travelling wave $|u_t|\propto|u_x|$), and the coordinates $x$ and $t$
are used only as controls. For a single Schr\"odinger soliton the phase gradient $\partial_x\arg\psi$ equals the
soliton speed everywhere, so it is not a spatial concept and is not used.

\paragraph{Why both signed and unsigned fields?}
SAE codes are non-negative, so a single atom can follow a non-negative field such as $|u^*_x|$ but only one sign of a
signed field such as $u^*$. Signed fields therefore test whether the SAE splits a variable across a pair of atoms
(\S\ref{sec:bilateral_detail}).

\begin{table}[t]
\small\centering
\caption{Concept panel, computed from reference solutions with central differences on the evaluation grid.
Operator terms are signed; when an atom is aligned with an operator term we use its absolute value, because atom
codes are non-negative. For Schr\"odinger, $\psi$ is the complex field.}
\label{tab:concepts}
\setlength{\tabcolsep}{4pt}
\begin{tabular}{lll}
\toprule
PDE & Concepts & Operator terms \\
\midrule
Heat & $|u|$, $|u_x|$, $|u_{xx}|$ & $u_{xx}$ \\
Burgers (all variants) & $|u|$, $|u_x|$, $|u_{xx}|$, $|u_t|$, $u$ & $uu_x$ (or $u^2u_x$), $\nu u_{xx}$ \\
Allen--Cahn & $|u|$, interface ind.\ $e^{-u^2/0.05}$, interface grad.\ $|u_x|$, $|u^3-u|$, $u$ & $\epsilon^2u_{xx}$, $5u-5u^3$ \\
Convection & $|u|$, $|u_x|$, $u$, characteristic $x-\beta t$ & $\beta u_x$ \\
Schr\"odinger & $|\psi|$, $\mathrm{Re}\,\psi$, $\mathrm{Im}\,\psi$, $|\partial_x|\psi||$ & $\tfrac12|\psi_{xx}|$, $|\psi|^3$ \\
KdV & $|u|$, $|u_x|$, $|u_{xx}|$, $|u_{xxx}|$ & $6uu_x$, $u_{xxx}$ \\
Fisher--KPP & $|u|$, $|u_x|$, $|u_{xx}|$, $u(1-u)$, $u-\tfrac12$ & $Du_{xx}$, $ru(1-u)$ \\
Advection--diffusion & $|u|$, $|u_x|$, $|u_{xx}|$, $|u_t|$ & $au_x$, $Du_{xx}$ \\
\bottomrule
\end{tabular}
\end{table}

\subsection{Sparse autoencoder}
\label{sec:sae}

\paragraph{Architecture.}
The SAE maps a z-scored hidden state $\tilde h\in\mathbb{R}^{d_h}$ to a non-negative code $z\in\mathbb{R}^D$ and back:
\begin{align}
  z(\tilde{h}) &= \mathrm{ReLU}\!\left(W_{\mathrm{enc}}\,\tilde{h} + b_{\mathrm{enc}}\right),
  \label{eq:sae_encode}\\
  \hat{\tilde{h}}(z) &= W_{\mathrm{dec}}\,z + b_{\mathrm{dec}} = \textstyle\sum_{k=1}^{D} z_k d_k + b_{\mathrm{dec}},
  \label{eq:sae_decode}\\
  \text{s.t.}\quad &\|d_k\|_2 = 1 \quad \forall k,
  \label{eq:norm_constraint}
\end{align}
where $W_{\mathrm{enc}} \in \mathbb{R}^{D \times d_h}$, $W_{\mathrm{dec}} \in \mathbb{R}^{d_h \times D}$, and $d_k$, the
$k$-th column of $W_{\mathrm{dec}}$, is the \emph{decoder direction} of atom $k$. The code $z_k(x,t)$ says how strongly
atom $k$ is present at the point $(x,t)$; viewed over the whole grid, $z_k$ is the atom's \emph{activation map}. Because
$\|d_k\|=1$, all scale information sits in $z_k$, so a larger code unambiguously means a stronger presence of the
direction.

We use $D=512$ atoms for $d_h=128$ units, an overcompleteness of $4\times$. With fewer atoms than units the SAE would
compress the layer; with more, it can assign separate atoms to features that share units.

\paragraph{Training objective.}
\begin{equation}
  \mathcal{L}_{\mathrm{SAE}}
  = \underbrace{\mathbb{E}_{\tilde{h}}\!\left[\|\tilde{h} - \hat{\tilde{h}}\|_2^2\right]}_{\text{reconstruction}}
  + \lambda\,\underbrace{\mathbb{E}_{\tilde{h}}\!\left[\|z(\tilde{h})\|_1\right]}_{\text{sparsity}},
  \qquad \lambda = 2\times10^{-2}.
  \label{eq:sae_loss}
\end{equation}
The reconstruction term keeps the code faithful to the PINN's hidden state; the L1 term pushes most codes to zero,
so that each atom is active only in part of the domain. With this setting about 170 of the 512 atoms are active at a
typical point, and the fraction of variance left unexplained on held-out points is below $10^{-3}$.

\paragraph{Unit-norm projection.}
After every Adam step the decoder columns are projected back to unit norm, $d_k \leftarrow d_k/\|d_k\|_2$. Without
this the scale could move freely between encoder, decoder and codes. The decoder is initialised as the transpose of
the encoder (tied initialisation), and training uses Adam with learning rate $10^{-3}$, batch size 8192 and 600 epochs
on the training split. We train three SAEs with different seeds for each PINN.

\paragraph{No pre-encoder bias.}
Some SAEs subtract a learned bias before the encoder to absorb the activation mean. Z-scoring already removes the
mean of every unit, so we omit it.

\paragraph{Variants.}
Section~\ref{sec:ablations} compares this default with smaller dictionaries ($D=64$), TopK SAEs~\citep{gao2024topk},
which keep exactly the $k$ largest pre-activations instead of using an L1 penalty, and derivative-augmented (``jet'')
SAEs whose input is $[\tilde h,\tilde h_x,\tilde h_t,\tilde h_{xx}]$, with derivatives obtained by forward-mode
differentiation.

\subsection{Concept alignment and linear probes}
\label{sec:alignment_method}

\paragraph{Alignment matrix.}
For each atom $k$ and concept $C_j$ we compute the Pearson correlation over the $N_{\mathrm{eval}}$ held-out points,
\begin{equation}
  A_{kj} = \mathrm{corr}(z_k, C_j)
  = \frac{\sum_{(x,t)} (z_k(x,t)-\bar{z}_k)(C_j(x,t)-\bar{C}_j)}
         {\sqrt{\sum_{(x,t)}(z_k-\bar{z}_k)^2}\,\sqrt{\sum_{(x,t)}(C_j-\bar{C}_j)^2}}.
  \label{eq:alignment}
\end{equation}
$A_{kj}$ measures how closely the activation map of atom $k$ follows concept $j$ in space and time. The best-aligned
atom for concept $j$ is
\begin{equation}
  k^\star_j = \operatornamewithlimits{arg\,max}_{k} |A_{kj}|.
  \label{eq:top_atom}
\end{equation}
We use $|A_{kj}|$ because an atom that is active exactly where a concept is small carries the same information as one
that is active where it is large.

\paragraph{Linear probes.}
A single atom can capture only part of what a layer holds. To measure how much of a concept is linearly decodable
from the layer as a whole, independently of the SAE, we fit a \emph{linear probe}: a ridge regression from
$\tilde h_\ell$ to $C_j$ on the training points, scored by $R^2$ on the held-out points. The ridge penalty is chosen by
generalised cross-validation~\citep{golub1979gcv} over $10^{-3}$ to $10^{3}$. A probe $R^2$ near 1 means the concept is
an almost exact linear function of the hidden state.

\subsection{Null calibration}
\label{sec:nulls}

A high correlation or probe $R^2$ shows that a concept is present, but not that training put it there: a network of
the same shape with random weights also computes many smooth functions of $(x,t)$. We therefore build, for each PDE,
a set of \emph{null networks}: 5 untrained networks with the same architecture, and the PINNs trained on the other
development PDEs (15 networks for the development PDEs, all 18 development PINNs for the held-out PDEs). Each null
network is evaluated on the same grid, with $x$ mapped linearly into its own training domain, and put through the
identical pipeline (z-scoring, its own SAE, probes, alignment). For every statistic we take the 95th percentile of
its null distribution, $\pnf$. A trained PINN \emph{beats the null} on a statistic when its value exceeds $\pnf$.

A permutation test, which shuffles a concept across grid points, is not a useful null here. Shuffling destroys
spatial smoothness, so every smooth concept passes it easily (permutation $Z>5$ for all concepts in our data).

\subsection{Direct causal intervention in the original hidden state}
\label{sec:intervention}

Alignment shows that an atom's activation map follows a concept. To test whether the atom matters for the PINN's
computation, we remove it from the PINN's own hidden state, bypassing the SAE decoder, and measure what changes.

\paragraph{Intervention formulation.}
The SAE works in z-scored coordinates, so the contribution of atom $k$ to the raw hidden state of layer $\ell$ at
$(x,t)$ is
\begin{equation}
  \text{contribution}_k(x,t) = z_k(x,t)\,(\sigma_\ell \odot d_k) \in \mathbb{R}^{d_h},
  \label{eq:contribution}
\end{equation}
where $\odot$ is the element-wise product and $\sigma_\ell$ the vector of standard deviations from Eq.~\eqref{eq:zscore}.
The counterfactual hidden state obtained by removing atom $k$ with strength $\alpha$ is
\begin{equation}
  h_\ell^{\mathrm{cf}}(x,t;\alpha) = h_\ell(x,t) - \alpha\,z_k(x,t)\,(\sigma_\ell\odot d_k).
  \label{eq:cf_hidden}
\end{equation}
At $\alpha=1$ this removes the atom's contribution exactly; $\alpha\in(0,1)$ suppresses it partially and $\alpha>1$
over-removes it. The counterfactual state is propagated through the remaining layers to give $u_{\mathrm{cf}}$, and
the effect of the intervention is $\Delta u(x,t) = u_\theta(x,t) - u_{\mathrm{cf}}(x,t)$. Since the decoder bias
$b_{\mathrm{dec}}$ and the reconstruction error are left in place, the intervention isolates the term $z_kd_k$ of the
decomposition $\tilde h = \sum_k z_kd_k + b_{\mathrm{dec}} + \varepsilon$ and does not depend on how well the SAE
reconstructs the layer.

\paragraph{Choice of layer.}
At the last hidden layer the output is linear in $h_L$, so
\begin{equation}
  \Delta u(x,t;\alpha) = \alpha\,z_k(x,t)\,W_{\mathrm{out}}(\sigma_L\odot d_k).
  \label{eq:delta_u}
\end{equation}
Here the intervention map $|\Delta u|$ is simply the atom's activation map times a constant (their correlation is
$1.000000$ in our runs), and its shape does not depend on $\alpha$. Intervening at layer 5 therefore measures where
the atom is active. At layers 3 and 4 the change passes through nonlinear layers before reaching the output, the
correlation between $|\Delta u|$ and $z_k$ drops to 0.76--0.91, and the footprint becomes an independent measurement.
We report footprints at both kinds of layer and use layer 4 for the residual analysis.

\paragraph{Causal metrics.}
We summarise the spatial distribution of $|\Delta u|$ over the held-out points by:
\begin{enumerate}[leftmargin=*,itemsep=2pt]
\item \textbf{Effective spatial footprint, $\ESF$.} Sort the points by $|\Delta u|$ in decreasing order and take
the smallest fraction of them that together carry 80\% of the total $\sum|\Delta u|$:
\begin{equation}
  \mathrm{ESF}_\delta = \frac{1}{N}\,\min\Bigl\{m : \textstyle\sum_{i=1}^{m} |\Delta u|_{(i)} \;\geq\; \delta \sum_{i=1}^{N} |\Delta u|_{(i)}\Bigr\},
  \qquad \delta = 0.8,
  \label{eq:esf}
\end{equation}
where $|\Delta u|_{(1)}\geq|\Delta u|_{(2)}\geq\dots$ are the sorted values. A small $\ESF$ means the effect is
concentrated in a small part of the domain; a uniform effect gives $\ESF=0.8$.
\item \textbf{Support capture.} Let $\mathcal{M}_j$ be the 20\% of points where $|C_j|$ is largest. Support capture is
the share of the effect that falls inside this region,
\begin{equation}
  \mathrm{SC}_j = \frac{\sum_{(x,t)\in\mathcal{M}_j} |\Delta u(x,t)|}{\sum_{(x,t)} |\Delta u(x,t)|}.
  \label{eq:support_capture}
\end{equation}
An effect spread evenly over the domain gives $\mathrm{SC}=0.2$; a value well above 0.2 means the effect is
concentrated where the concept is large.
\item \textbf{Residual change.} The change in the mean squared PDE residual,
$\log_{10}\bigl(\overline{\|\mathcal{F}[u_{\mathrm{cf}}]\|^2}/\overline{\|\mathcal{F}[u_\theta]\|^2}\bigr)$, which asks
whether the removed atom matters for satisfying the equation rather than only for the value of $u$.
\end{enumerate}

\paragraph{Baseline decompositions.}
We apply the same intervention to the components of three other linear decompositions of $\tilde h$, so that the
only difference between methods is the decomposition: PCA (principal component $v_k$ with score
$a_k=\tilde h^\top v_k$, removed as $a_k(\sigma\odot v_k)$), ICA (FastICA~\citep{hyvarinen1999ica}, source $s_k$ with
unit-norm mixing vector $m_k$, removed as $s_k(\sigma\odot m_k)$), and a \emph{sparse linear dictionary}, which is the
same autoencoder with the ReLU removed, so that codes can be negative.

\subsection{Matched control atoms}
\label{sec:neg_controls_method}

To check that an effect is specific to concept-aligned atoms rather than a property of removing any active atom, each
top atom $k^\star_j$ is compared with 10 control atoms. We use two matching rules. \emph{Activation-matched}
controls are the 10 atoms whose mean activation $\bar z_k$ is closest to that of $k^\star_j$. (Matching on decoder norm
as well would add nothing, since all decoder columns have unit norm.) \emph{Output-matched} controls are atoms that are
not aligned with the concept ($|A_{kj}|<0.3$) and whose root-mean-square output change $\|\Delta u\|_{\mathrm{rms}}$ is
closest to that of the top atom. Output matching guarantees that a larger effect on the residual or on the concept
region cannot come from a larger effect on the output overall.

For footprints we report the \textbf{control gap}
\begin{equation}
  \Delta_{\mathrm{ctrl}} = \mathrm{mean}_{k\in\mathrm{ctrl}}\bigl[\ESF^{(k)}\bigr] - \ESF^{(k^\star_j)}.
  \label{eq:advantage}
\end{equation}
With this sign convention a \emph{positive} gap means the top atom's effect is more concentrated than that of its
controls.

\subsection{Bilateral and multi-atom representations}
\label{sec:bilateral_detail}

ReLU codes are non-negative, so a signed concept such as $u^*$ needs at least two atoms: one active where $u^*>0$ and
one where $u^*<0$. We call this a \emph{bilateral} representation.

\paragraph{Detection and scoring.}
On the training points we find the top atom $k^+=\arg\max_k A_{kj}$ and the most anti-correlated atom
$k^-=\arg\min_k A_{kj}$, fit ordinary least squares models $C_j\sim z_{k^+}$ and $C_j\sim z_{k^+}+z_{k^-}$, and report
$R^2$ on the held-out points. Adding a regressor can never lower the in-sample $R^2$, so a fair comparison needs a
partner of the same complexity: we also fit $k^+$ together with the best same-sign atom, and with each of 20 random
atoms. A bilateral code shows up as a clearly larger held-out $R^2$ for the opposite-sign partner than for the others.
The \emph{two-atom gain} is the $R^2$ with the opposite-sign partner minus that with a random partner, and is compared
with the same gain on null networks.

\subsection{Distinguishing atoms}
\label{sec:distinct_method}

Activation maps of different atoms in the same PINN can look alike, either because the SAE has split one feature
across several atoms or because the concepts themselves overlap in space. We use three checks to decide whether two
atoms are different features.
\begin{enumerate}[leftmargin=*,itemsep=2pt]
\item \textbf{Distinct assignment.} Choosing $k^\star_j$ independently for each concept can select the same atom
twice. For galleries we instead assign one distinct atom to each concept by maximising the total $|A_{kj}|$ with the
Hungarian algorithm~\citep{kuhn1955hungarian}, which solves this one-to-one matching problem exactly.
\item \textbf{Selectivity.} For an assigned atom $k_j$, its selectivity is
$|A_{k_j j}| - \max_{j'\neq j}|A_{k_j j'}|$, the margin by which it prefers its own concept to every other concept in
the panel, and its partial correlation is the correlation with $C_j$ after linearly regressing out the other concepts.
Positive selectivity and a high partial correlation mean the atom carries something specific to its concept.
\item \textbf{Concept overlap.} We compare the correlation between assigned atoms with the correlation between their
concepts. If two concepts are large in the same places (the amplitude and the amplitude gradient of a soliton, say),
atoms that track them must also be correlated, and similarity of their maps is expected rather than a defect.
\end{enumerate}
One-dimensional profiles at a fixed time make the remaining differences visible (Figure~\ref{fig:distinct}).

\subsection{Feature consistency across SAE seeds}
\label{sec:consistency_method}

Sparse dictionaries are not unique: different dictionaries can reach the same reconstruction and sparsity. For three
SAE seeds per PINN we match decoder columns between every pair of seeds with the Hungarian algorithm, maximising the
total cosine similarity
\begin{equation}
  \pi^* = \operatornamewithlimits{arg\,max}_{\pi \in \mathcal{S}_D}
  \sum_{k=1}^D \frac{(d_k^{(s)})^\top d_{\pi(k)}^{(s')}}{\|d_k^{(s)}\|\,\|d_{\pi(k)}^{(s')}\|},
  \label{eq:hungarian}
\end{equation}
and report the mean matched cosine. As a reference we compute the same quantity for two random Gaussian
dictionaries of the same size. Because decoder directions and activation maps can behave differently, we also report
the correlation between the activation maps of the best-aligned atoms in different seeds.

\subsection{Pre-registered confirmation}
\label{sec:prereg_method}

Analyses that are tuned while looking at the data can overstate their findings. To guard against this, the main
claims were tested on held-out PDEs that had not been used during development~\citep{nosek2018prereg}. The
predictions, thresholds and scoring rules were written into the analysis code, and a timestamped record with a hash of
that code was stored before any held-out PINN was trained. Scoring is automatic.

\subsection{Algorithm summary}
\label{sec:algorithm}

\hyperref[alg:physsae]{Algorithm~\ref*{alg:physsae}} summarises the procedure for one PINN.

\begin{algorithm}[h]
\caption{\physsae{} for one trained PINN}
\label{alg:physsae}
\begin{algorithmic}[1]
\Require Trained PINN $u_\theta$; reference $u^*$; grid $\mathcal{G}$; sparsity $\lambda$; dictionary size $D$; null networks.
\Statex \textbf{--- Stage 1: activations ---}
\State Evaluate the PINN on $\mathcal{G}$; record $H_\ell$ for every layer; z-score each unit, keep $(\mu_\ell,\sigma_\ell)$
\State Split grid points 80/20 into training and held-out sets
\Statex \textbf{--- Stage 2: dictionaries and probes ---}
\For{SAE seed $s = 1, 2, 3$}
  \State Train the SAE on the training points (Eq.~\eqref{eq:sae_loss}), projecting $d_k$ to unit norm after every step
\EndFor
\State Fit ridge probes from $\tilde h_\ell$ to every concept; fit PCA, ICA and the sparse linear dictionary
\State Match decoder columns across seeds (Eq.~\eqref{eq:hungarian})
\Statex \textbf{--- Stage 3: alignment ---}
\State Compute concept fields $\{C_j\}$ from $u^*$; encode held-out activations; compute $A$ (Eq.~\eqref{eq:alignment})
\State Select $k^\star_j$ (Eq.~\eqref{eq:top_atom}) and distinct assignments; compute selectivity
\Statex \textbf{--- Stage 4: intervention ---}
\For{each concept $j$ and layer $\ell$}
  \State $h_\ell^{\mathrm{cf}} = h_\ell - \alpha\,z_{k^\star_j}(\sigma_\ell\odot d_{k^\star_j})$; propagate to $u_{\mathrm{cf}}$; compute $\Delta u$ and $\mathcal{F}[u_{\mathrm{cf}}]$
  \State Compute $\ESF$, support capture and residual change; repeat for matched controls and baseline components
\EndFor
\Statex \textbf{--- Stage 5: calibration ---}
\State Repeat Stages 1--3 on every null network; compare each statistic with its null $\pnf$
\State Fit bilateral pairs and score them on held-out points
\end{algorithmic}
\end{algorithm}

\section{Experimental setup}
\label{sec:setup}

\subsection{PDE benchmark}
Table~\ref{tab:pdes} lists the 12 PDEs. The six \emph{development} PDEs were used to build and tune the analysis.
They deliberately span converged (heat, Schr\"odinger, convection with $\beta=1$), partially converged (Burgers) and
known-failure cases (Allen--Cahn, convection with $\beta=30$). The six \emph{held-out} PDEs were used only
after the analysis was fixed. They include two travelling waves (the KdV soliton and the Fisher--KPP front), a sharper
Burgers shock, and three equations in which sharp features form or spread instead of travelling rigidly (Burgers and
modified Burgers from a Gaussian, and advection--diffusion). The last three carry the operator-term test: for a rigidly
travelling wave $u_x$ is a function of $u$ alone, which would make a test of derivative encoding nearly circular. A PDE
counts as \emph{well trained} if its mean relative $L^2$ error over three seeds is below 0.20.

\begin{table}[t]
\small\centering
\caption{PDE benchmark. Relative $L^2$ error of the PINN against the reference, mean over 3 seeds. All domains are
$[-1,1]\times[0,1]$ except Schr\"odinger and KdV ($[-5,5]\times[0,1]$).}
\label{tab:pdes}
\setlength{\tabcolsep}{3pt}
\begin{tabular}{lllll}
\toprule
& PDE & Equation & Initial condition & $L^2$ err \\
\midrule
\multirow{6}{*}{\rotatebox{90}{development}}
 & Heat           & $u_t = u_{xx}$                                    & $e^{-5x^2}$        & 0.001 \\
 & Burgers        & $u_t+uu_x=\nu u_{xx}$, $\nu{=}0.01/\pi$            & $-\sin(\pi x)$     & 0.126 \\
 & Allen--Cahn    & $u_t=10^{-4}u_{xx}+5u-5u^3$                       & $x^2\cos(\pi x)$   & 0.934 \\
 & Convection     & $u_t+u_x=0$                                       & $\sin(\pi x)$      & 0.032 \\
 & Convection     & $u_t+30u_x=0$                                     & $\sin(\pi x)$      & 0.984 \\
 & Schr\"odinger  & $i\psi_t+\tfrac{1}{2}\psi_{xx}+|\psi|^2\psi=0$    & soliton, amplitude 2, speed 2 & 0.023 \\
\midrule
\multirow{6}{*}{\rotatebox{90}{held-out}}
 & KdV            & $u_t+6uu_x+u_{xxx}=0$                             & soliton, $c=2$     & 0.002 \\
 & Fisher--KPP    & $u_t=0.01u_{xx}+5u(1-u)$                          & travelling front   & 0.002 \\
 & Burgers, sharp & $u_t+uu_x=\nu u_{xx}$, $\nu{=}0.003/\pi$           & $-\sin(\pi x)$     & 0.302 \\
 & Burgers, Gaussian & $u_t+uu_x=\nu u_{xx}$, $\nu{=}0.01/\pi$         & $e^{-25(x+0.4)^2}$ & 0.014 \\
 & Modified Burgers  & $u_t+u^2u_x=\nu u_{xx}$, $\nu{=}0.02/\pi$       & $e^{-25(x+0.4)^2}$ & 0.012 \\
 & Advection--diffusion & $u_t+0.5u_x=0.002u_{xx}$                    & erfc front         & 0.002 \\
\bottomrule
\end{tabular}
\end{table}

\subsection{Hyperparameters and compute}
Full settings are listed in Appendix~\ref{app:hyperparams}. All experiments ran on a single NVIDIA RTX 4090 (24\,GB)
and took about three and a half hours in total. Code, result files and the pre-registration record are released
with the paper.

\section{Results}
\label{sec:results}

\subsection{PINN training quality}

Nine of the twelve PDEs train to relative errors between 0.1\% and 13\% (Table~\ref{tab:pdes}). Allen--Cahn and
convection with $\beta=30$ fail (errors 0.93 and 0.98), in line with the known difficulty of stiff reaction and
high-speed advection for PINNs~\citep{krishnapriyan2021failures}, and the sharp Burgers problem is intermediate (0.30).
We keep the failed PINNs on purpose: they show what \physsae{} reports when a network has not learned its equation
(\S\ref{sec:collapse}).

\subsection{SAE atoms align with physical fields}
\label{sec:alignment_results}

\begin{figure}[t]
\centering
\begin{subfigure}[t]{0.49\linewidth}\centering
\includegraphics[width=\linewidth]{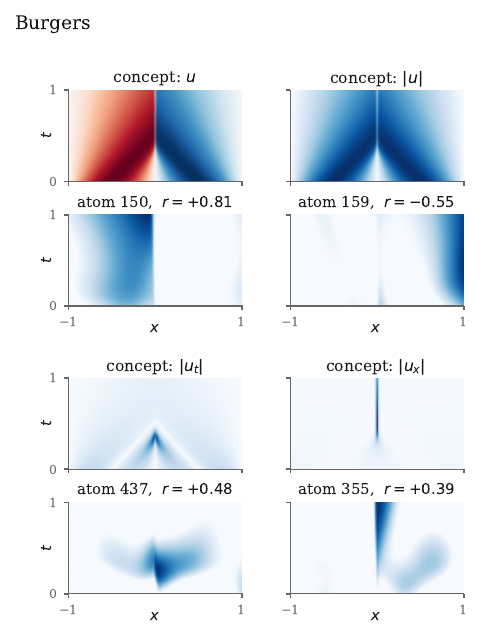}
\caption{Burgers}
\end{subfigure}\hfill
\begin{subfigure}[t]{0.49\linewidth}\centering
\includegraphics[width=\linewidth]{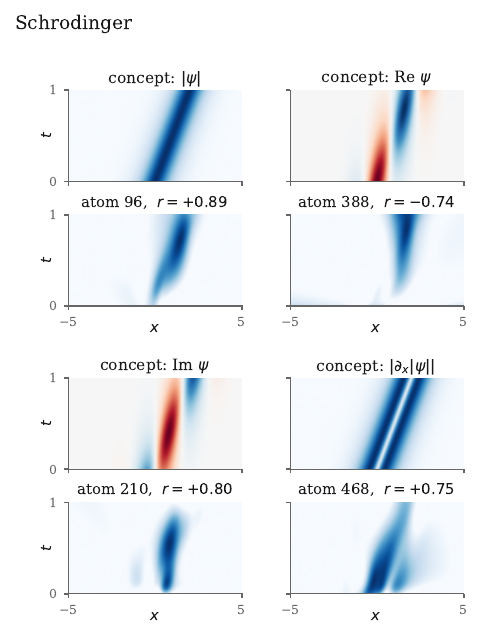}
\caption{Schr\"odinger}
\end{subfigure}
\caption{SAE atoms against independently computed concept fields (layer 5, PINN seed 0, SAE seed 0). Each pair
of panels shows a concept from the reference solution (upper) and the activation map of the atom assigned to it
(lower; $x$ horizontal, $t$ vertical), with their correlation on held-out points. Atoms are assigned one-to-one by the
Hungarian algorithm, so no atom appears twice. The atoms were learned from hidden activations alone.}
\label{fig:atoms}
\end{figure}

\begin{figure}[t]
\centering
\includegraphics[width=0.95\linewidth]{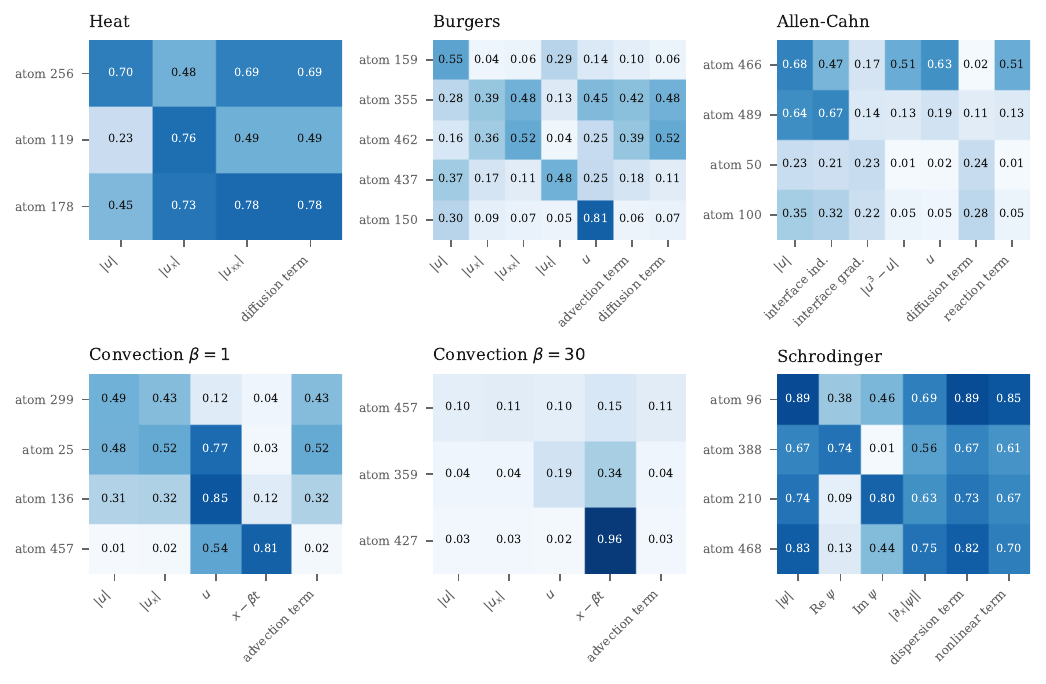}
\caption{Alignment $|A_{kj}|$ between the best-aligned atoms (rows) and all concepts (columns) for the six development
PDEs (layer 5, PINN seed 0, SAE seed 0). Most atoms align strongly with one or two concepts.}
\label{fig:heatmaps}
\end{figure}

On well-trained PINNs the SAE atoms form spatially coherent features that follow the physics
(Figure~\ref{fig:atoms}). In the Burgers PINN, separate atoms follow the signed solution, its magnitude, its rate of
change, and the shock. In the Schr\"odinger PINN, atoms follow the moving soliton,
its real and imaginary parts, which oscillate inside the envelope, and the two flanks where the amplitude changes
fastest. The atoms were never shown these fields: they were learned from the hidden activations alone and only
compared with the reference afterwards. Table~\ref{tab:alignment} summarises the strongest alignment per PDE, and
Figure~\ref{fig:heatmaps} shows how each of the best atoms relates to every concept. Most atoms align strongly with one
or two concepts and weakly with the rest. The remaining galleries are in Appendix~\ref{app:atoms}.

\begin{table}[t]
\small\centering
\caption{Strongest alignment per development PDE (layer 5). $|r|$: best-atom correlation for the best-aligned
concept, mean over 3 PINN $\times$ 3 SAE seeds. Last two columns: number of concepts for which the trained PINN beats
the null 95th percentile, for the best SAE atom and for a linear probe on the full layer (\S\ref{sec:null_results}).}
\label{tab:alignment}
\setlength{\tabcolsep}{5pt}
\begin{tabular}{llccc}
\toprule
PDE & Best-aligned concept & $|r|$ & SAE $>$ null & Probe $>$ null \\
\midrule
Heat          & curvature $|u_{xx}|$           & 0.800 & 0/3 & 1/3 \\
Burgers       & signed value $u$               & 0.800 & 2/5 & 5/5 \\
Allen--Cahn   & interface indicator            & 0.692 & 1/5 & 0/5 \\
Convection $\beta=1$  & characteristic $x-t$   & 0.825 & 3/4 & 3/4 \\
Convection $\beta=30$ & characteristic $x-30t$ & 0.941 & 4/4 & 3/4 \\
Schr\"odinger & amplitude $|\psi|$             & 0.848 & 3/4 & 4/4 \\
\midrule
Total & & & 13/25 & 16/25 \\
\bottomrule
\end{tabular}
\end{table}

\subsection{Telling atoms apart}
\label{sec:distinct_results}

\begin{figure}[t]
\centering
\includegraphics[width=\linewidth]{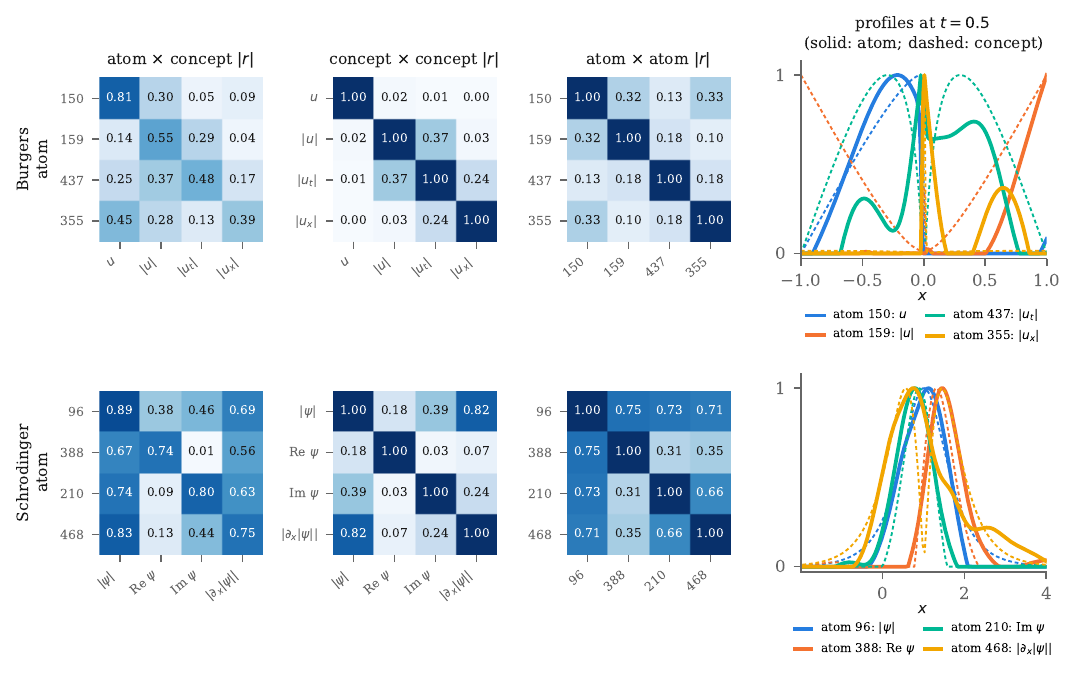}
\caption{Telling atoms apart (layer 5, PINN seed 0, SAE seed 0; atoms as in Figure~\ref{fig:atoms}). Left to right:
correlation of each assigned atom with each concept (a dominant diagonal means each atom prefers its own concept);
correlation among the concepts themselves; correlation among the atoms; and profiles at $t=0.5$ of each atom (solid)
and of the part of its concept that it can follow (dashed; for an anti-correlated atom the complement of the concept).}
\label{fig:distinct}
\end{figure}

Several atoms in Figure~\ref{fig:atoms} have activation maps that look alike, particularly in the Schr\"odinger PINN,
where every panel is a band along the soliton's path. This raises a fair question: are these different features, or
one feature seen several times? Figure~\ref{fig:distinct} applies the three checks of \S\ref{sec:distinct_method}.

In the Burgers PINN the checks separate the atoms well. The concepts are nearly uncorrelated with each other
(mean $|r|=0.11$), the assigned atoms only weakly (0.21), and the atoms for $u$, $|u|$ and $|u_t|$ each prefer their own
concept on every seed, with partial correlations of 0.78--0.86 for $u$. The maps and profiles show why: the $u$ atom covers the
positive half of the solution, the $|u|$ atom (anti-correlated with $|u|$) fires where $|u|$ is small, at the right
boundary and along the shock line, and the $|u_t|$ atom follows the region around the forming shock where the solution
changes fastest in time. The shock-gradient atom is the weak one. It
correlates with $u$ slightly more than with $|u_x|$ (selectivity $-0.01$ to $-0.05$), so the shock is shared with the
signed-value atoms, which matches the probe result that the gradient is carried by many atoms together
(\S\ref{sec:ablations}).

In the Schr\"odinger PINN the four atoms share a common part. They correlate with each other at 0.59 on average, more
than their concepts do (0.29), because every one of them is active on the soliton envelope. What tells them apart is what
remains after this shared part is removed: the partial correlation with the own concept, after regressing out the other
three, is 0.76--0.91 for $|\psi|$, 0.79--0.88 for $\mathrm{Re}\,\psi$ and 0.73--0.91 for $\mathrm{Im}\,\psi$. These three
atoms therefore carry information specific to their concept, and their profiles peak at different positions inside the
pulse. The amplitude-gradient atom does not pass this test (partial 0.21--0.35): it is mostly the envelope. Across the
development PDEs the own concept is the best match for its assigned atom in 46 of 69 cases, and in 45 of 57 cases on the
well-trained PDEs (Appendix Table~\ref{tab:app_distinct}).

The practical rule we draw is that similar-looking atoms should be compared through their selectivity and partial
correlation, not their maps alone. Atoms that track quantities living in the same place will look alike, and the partial
correlation shows whether each still carries something of its own.

\subsection{Which alignments come from training}
\label{sec:null_results}

\begin{figure}[t]
\centering
\includegraphics[width=\linewidth]{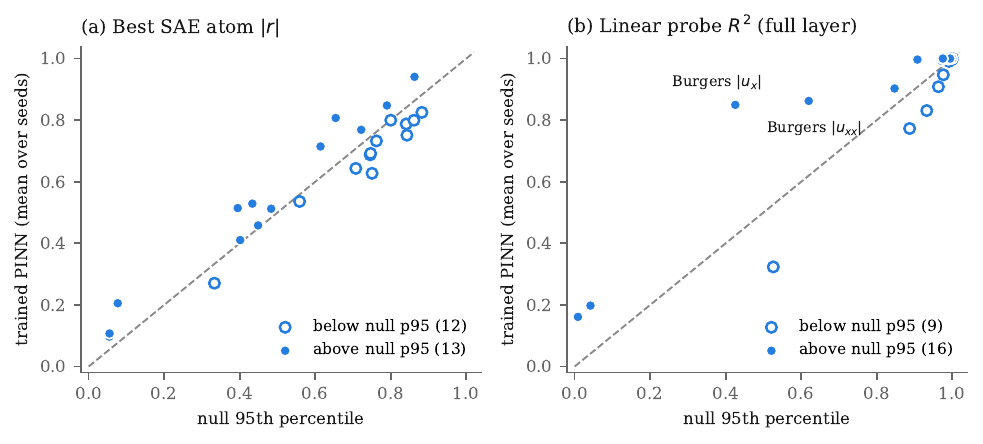}
\caption{Trained PINNs against the 95th percentile of their null networks (5 untrained, 15 trained on other PDEs) for
the 25 development concepts. (a) Best SAE atom $|r|$, mean over 3 PINN $\times$ 3 SAE seeds. (b) Linear probe $R^2$ on
the full layer, mean over 3 PINN seeds. Points above the diagonal beat the null. Values in Appendix
Table~\ref{tab:app_null}.}
\label{fig:nullcal}
\end{figure}

Figure~\ref{fig:nullcal} compares each of the 25 development concepts with its null distribution. The best SAE atom
beats the null for 13 concepts and a linear probe on the full layer for 16. The concepts that beat the null are
exactly those where the solution has structure that a generic network does not produce. The Burgers shock is the
clearest case: a probe recovers $|u_x|$ with $R^2=0.850$ against a null $\pnf$ of 0.425, and $|u_{xx}|$ with 0.862
against 0.620. The oscillating solution of convection with $\beta=1$ and the Schr\"odinger soliton, including its real
and imaginary parts, also beat the null with both methods.

Smooth fields are where the null networks catch up. For the heat equation, untrained networks and PINNs trained on other
PDEs give atoms that correlate with $|u|$, $|u_x|$ and $|u_{xx}|$ of the heat solution as strongly as the heat PINN's own
atoms do, and probes reach $R^2\approx1$ on every network. This does not mean the heat PINN failed to learn its
solution (its error is 0.1\%); it means a smooth Gaussian is too easy a target to reveal anything specific to training.
This ceiling is why the pre-registered test below uses PDEs with sharp features and requires a fixed margin above the
null.

\subsection{PINNs encode the terms of their own operator}
\label{sec:operator}

The operator terms are the sharpest test of what a PINN has learned. The network is trained to make these terms cancel,
and nothing pushes a network trained on another equation, or an untrained one, to represent them.

\paragraph{Development.}
In the Burgers PINN a probe on layer 5 recovers the advection term $uu_x$ with $R^2=0.450$ and the diffusion term
$\nu u_{xx}$ with $R^2=0.657$, against null $\pnf$ of 0.074 and 0.072. At layer 3 the same probes give only 0.057 and
0.027, so the terms are assembled in the last two hidden layers. SAE atoms single out operator terms as well: in the
Schr\"odinger PINN one atom follows the nonlinear term with $|r|=0.83$ (null $\pnf$ 0.65), in KdV atoms follow both the
nonlinear and the dispersive term (0.81 and 0.83 against 0.72 and 0.71), and in Fisher--KPP an atom follows the reaction
term at 0.98 (null 0.89). Over the development PDEs, KdV and Fisher--KPP, operator terms beat the null in 11 of 15 cases (Appendix
Table~\ref{tab:app_terms_dev}).

\begin{figure}[t]
\centering
\includegraphics[width=\linewidth]{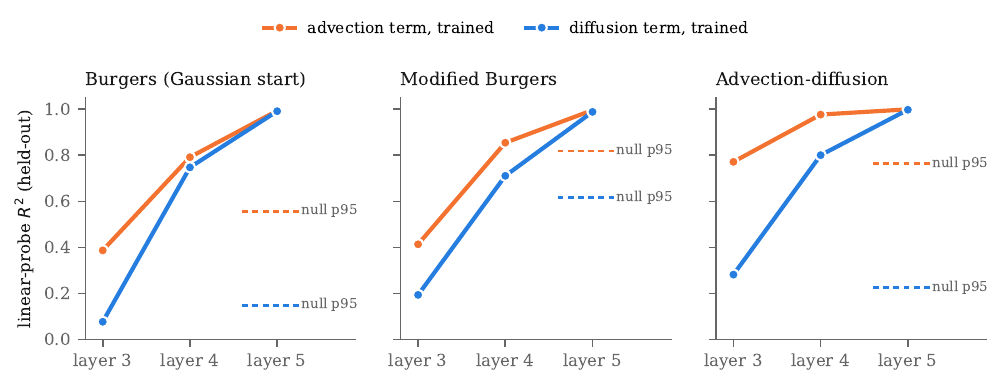}
\caption{Pre-registered confirmation: probe $R^2$ for the advection and diffusion terms at layers 3--5 of the trained
PINNs (mean of 3 seeds), with the 95th percentile of 23 null networks at layer 5 (dashed).}
\label{fig:depth}
\end{figure}

\paragraph{Pre-registered confirmation.}
We then fixed five predictions, with a margin of 0.05 in $R^2$ above the null, and trained PINNs on the three held-out
PDEs whose features form rather than travel. All nine PINNs reached errors of 0.2--1.7\%. All five predictions held, with 23 of 24 individual items passing
(Table~\ref{tab:phase6}, Figure~\ref{fig:depth}):
\begin{itemize}[leftmargin=*,itemsep=1pt]
\item every operator term is decodable from layer 5 well beyond the null (6/6), with the diffusion terms showing the
largest gaps (Gaussian Burgers: 0.991 against 0.149; advection--diffusion: 0.998 against 0.224);
\item decodability grows from layer 3 to layer 5 by more than it does in null networks (5/6);
\item SAE atoms align with every term beyond the null (6/6);
\item the derivative concepts $|u_x|$ and $|u_{xx}|$ beat the null on all three PDEs;
\item as a negative control, the amplitude $|u|$, which any network represents, does \emph{not} beat the null by the
margin on any PDE.
\end{itemize}
The one failed item is the depth gain of the advection term in the advection--diffusion equation, where layer 3 already
represents most of it.

\begin{table}[t]
\centering\small
\caption{Pre-registered confirmation on three held-out PDEs. Probe $R^2$ at layer 5 and its gain from layer 3 (mean of 3 seeds), and
best-atom $|r|$ with $|\text{term}|$ (mean of 3 PINN $\times$ 2 SAE seeds), each against the 95th percentile of 23 null
networks. Probe items need a margin of 0.05; a claim holds if at least 2/3 of its items pass. Bold: the one failed item.}
\label{tab:phase6}
\setlength{\tabcolsep}{4pt}
\begin{tabular}{llcccccc}
\toprule
& & \multicolumn{2}{c}{probe $R^2$} & \multicolumn{2}{c}{gain layer 3$\to$5} & \multicolumn{2}{c}{SAE $|r|$} \\
\cmidrule(lr){3-4}\cmidrule(lr){5-6}\cmidrule(lr){7-8}
PDE & term & PINN & null $\pnf$ & PINN & null $\pnf$ & PINN & null $\pnf$ \\
\midrule
Burgers (Gaussian) & $uu_x$ & 0.993 & 0.558 & 0.606 & 0.303 & 0.610 & 0.465 \\
 & $\nu u_{xx}$ & 0.991 & 0.149 & 0.914 & 0.087 & 0.682 & 0.520 \\
Modified Burgers & $u^2u_x$ & 0.996 & 0.819 & 0.582 & 0.412 & 0.692 & 0.660 \\
 & $\nu u_{xx}$ & 0.988 & 0.616 & 0.794 & 0.461 & 0.747 & 0.669 \\
Adv.--diff. & $a u_x$ & 0.999 & 0.763 & \textbf{0.228} & \textbf{0.297} & 0.863 & 0.570 \\
 & $D u_{xx}$ & 0.998 & 0.224 & 0.716 & 0.167 & 0.716 & 0.532 \\
\midrule
\multicolumn{2}{l}{Items passing} & \multicolumn{2}{c}{6/6} & \multicolumn{2}{c}{5/6} & \multicolumn{2}{c}{6/6} \\
\bottomrule
\end{tabular}
\end{table}

\paragraph{The diffusion terms are the strongest evidence.}
One might ask whether the hidden layer simply re-encodes the output $\hat u$, from which some terms could be computed. As
an exploratory check we predicted each term from a cubic polynomial in $(x,t,\hat u)$. For the advection terms this
baseline already reaches $R^2=0.84$, 0.81 and 0.96, so much of their decodability is explained by the output. For the
diffusion terms it reaches only 0.36, 0.58 and 0.70, against 0.99 from the hidden layer (Appendix
Table~\ref{tab:app_phase6}). The second-derivative terms are therefore represented in the hidden layer in a form the
output alone does not provide. (The polynomial has 20 features and the probe 128, so the comparison is indicative.)

\subsection{Direct causal intervention}
\label{sec:causal_results}

\begin{figure}[t]
\centering
\includegraphics[width=0.92\linewidth]{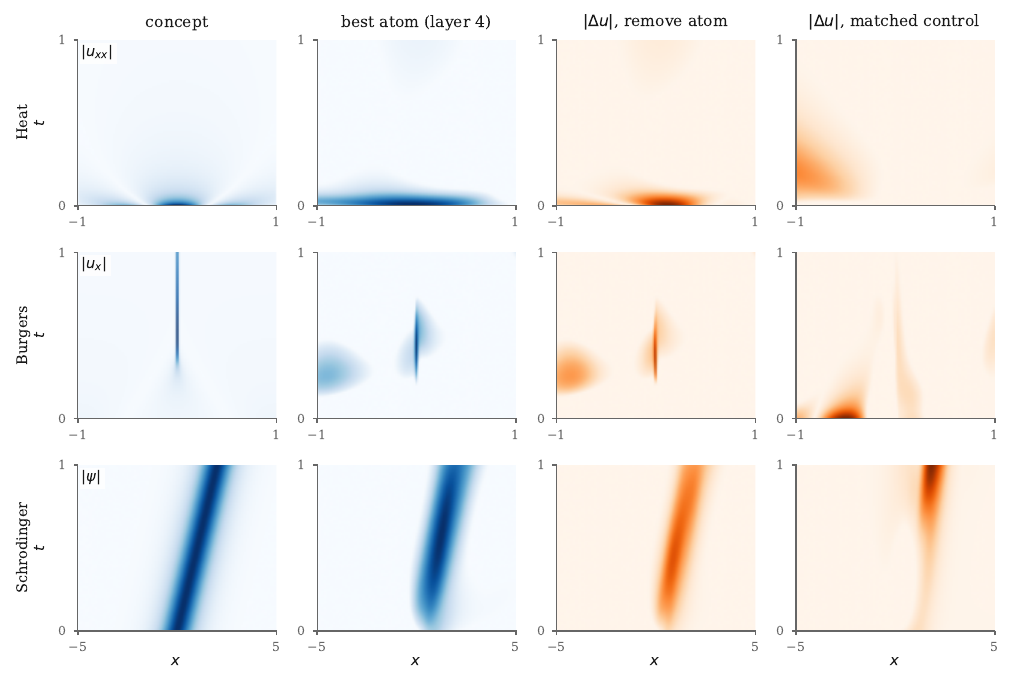}
\caption{Interventions at layer 4 (PINN seed 0, $\alpha=1$). Columns: concept field; activation of the best-aligned
atom; $|\Delta u|$ after removing that atom; $|\Delta u|$ after removing the output-matched control atom (same colour
scale within a row).}
\label{fig:interventions}
\end{figure}

Removing a concept-aligned atom from the PINN's hidden state changes the solution mainly where that concept is large
(Figure~\ref{fig:interventions}). Removing the atom aligned with the heat curvature, the Burgers shock gradient or the
Schr\"odinger amplitude at layer 4 concentrates $|\Delta u|$ on the region of the concept, while an output-matched
control atom, which changes the solution by the same overall amount, acts elsewhere. Averaged over concepts, top atoms place 48--49\% of
their output change inside the top-20\% region of their concept at each of layers 3, 4 and 5, against 21--26\% for
output-matched controls, 27--32\% for the best-aligned PCA component and 20\% for an effect spread uniformly. The top atom
beats its controls in all 18 combinations of PDE and layer (Appendix Table~\ref{tab:app_layers}). In the examples of
Figure~\ref{fig:interventions} the shares are 81\% against 26\% (heat), 52\% against 34\% (Burgers) and 94\% against
74\% (Schr\"odinger).

\paragraph{Comparison with PCA, ICA and a sparse linear dictionary.}
Table~\ref{tab:main} applies the same intervention to every decomposition at layer 5. SAE atoms have the most
concentrated footprints on all six PDEs: on average 0.179 of the domain carries 80\% of the effect, against 0.502 for PCA,
0.355 for ICA and 0.380 for the sparse linear dictionary, and the SAE footprint is 2.1--4.3$\times$ smaller than the PCA
one.
Because the layer-5 footprint of an atom equals its activation map up to a constant (Eq.~\eqref{eq:delta_u}), this
comparison measures how localised the codes of each decomposition are. At layer 4, where the footprint is an independent
measurement, SAE atoms remain more concentrated than PCA components ($\ESF=0.151$ against 0.387). Output-matched control
atoms from the same SAE are about as concentrated (0.140), so concentration is a property of the sparse code, while
concept specificity shows up in support capture.

\begin{table}[t]
\small\centering
\caption{Intervention footprints at layer 5 (mean over concepts; mean over 3 PINN seeds of the mean over 3 SAE seeds).
$\ESF$: fraction of held-out points carrying 80\% of $\sum|\Delta u|$ (lower is more concentrated). SD: sparse linear
dictionary. $\Delta_{\mathrm{ctrl}}$: control gap of Eq.~\eqref{eq:advantage} for activation-matched controls, all concepts
(positive means the top atom is more concentrated). SC: support capture on localised concepts for the
top atom and its controls (0.2 for a uniform effect).}
\label{tab:main}
\setlength{\tabcolsep}{3.5pt}
\begin{tabular}{lcccccccc}
\toprule
PDE & SAE $\ESF$ & PCA & ICA & SD & PCA/SAE & $\Delta_{\mathrm{ctrl}}$ & SC top & SC ctrl \\
\midrule
Heat           & 0.125 & 0.505 & 0.371 & 0.343 & 4.0 & $+0.051$ & 0.800 & 0.307 \\
Burgers        & 0.175 & 0.523 & 0.360 & 0.450 & 3.0 & $-0.008$ & 0.392 & 0.283 \\
Allen--Cahn    & 0.224 & 0.515 & 0.374 & 0.466 & 2.3 & $-0.052$ & 0.508 & 0.216 \\
Convection $\beta=1$  & 0.236 & 0.525 & 0.370 & 0.465 & 2.2 & $-0.060$ & 0.478 & 0.190 \\
Convection $\beta=30$ & 0.185 & 0.387 & 0.324 & 0.210 & 2.1 & $-0.086$ & 0.226 & 0.141 \\
Schr\"odinger  & 0.128 & 0.556 & 0.330 & 0.348 & 4.3 & $+0.059$ & 0.697 & 0.280 \\
\midrule
Mean           & \textbf{0.179} & 0.502 & 0.355 & 0.380 & 3.0 & $-0.016$ & \textbf{0.517} & 0.236 \\
\bottomrule
\end{tabular}
\end{table}

\paragraph{Effect on the PDE residual.}
Removing a concept-aligned atom also changes how well the PINN satisfies its equation, by a different amount than
removing an output-matched control (Table~\ref{tab:causal}). The clearest cases are at layer 4, where removing aligned
atoms raises the residual much more than controls in the heat and Schr\"odinger PINNs ($+0.34$ and $+0.54$ in
$\log_{10}$, factors of about 2 and 3.5), and in KdV. For the other PDEs the difference is small or reverses sign, most
strongly for Fisher--KPP. Aligned atoms are therefore tied to how the PINN satisfies its equation, but the size and sign of
the effect depend on the equation and the layer (\S\ref{sec:discussion}).

\begin{table}[t]
\centering\small
\caption{Change in PDE residual from removing the best-aligned atom minus that from removing output-matched controls
($\log_{10}$ ratio of mean squared residual after/before; L1 SAE; mean over concepts and 3 PINN seeds). Positive: the
aligned atom matters more for the residual.}
\label{tab:causal}
\begin{tabular}{lcc}
\toprule
PDE & layer 4 & layer 5 \\
\midrule
Heat & $+0.343$ & $-0.529$ \\
Burgers & $+0.009$ & $+0.093$ \\
Allen--Cahn & $-0.002$ & $+0.003$ \\
Convection $\beta=1$ & $-0.098$ & $-0.285$ \\
Convection $\beta=30$ & $-0.004$ & $-0.023$ \\
Schr\"odinger & $+0.537$ & $-0.064$ \\
\midrule
KdV (held-out; $|u_x|$, $|u_{xx}|$) & \multicolumn{2}{c}{$+0.034$ (layers 4 and 5)} \\
Fisher--KPP (held-out; $|u_x|$, $|u_{xx}|$) & \multicolumn{2}{c}{$-1.030$ (layers 4 and 5)} \\
\bottomrule
\end{tabular}
\end{table}

\subsection{Bilateral and multi-atom representations}
\label{sec:bilateral}

Signed fields are represented by pairs of atoms with opposite signs. For every signed concept, adding the most
anti-correlated atom to the top atom raises the held-out $R^2$ more than adding a same-sign atom or a random atom
(Table~\ref{tab:bilateral}). The Schr\"odinger real part is the clearest example: one atom explains $R^2=0.354$, the
opposite-sign pair 0.785, while a same-sign partner reaches 0.397 and a random partner 0.390. Part of this effect also
appears in null networks for smooth signed fields, because any ReLU dictionary has to split a signed variable across
two atoms. The trained PINNs exceed the null gain for the oscillating fields of convection with $\beta=1$ and of
Schr\"odinger, where the pairing reflects learned structure, but not for monotone fronts. That monotone fronts would not
exceed the null was itself predicted in advance and confirmed for Fisher--KPP.

\begin{table}[t]
\small\centering
\caption{Bilateral codes: held-out $R^2$ of the concept from the top atom alone and with one partner chosen on
training points (L1 SAE, mean of 3 PINN $\times$ 3 SAE seeds), and the two-atom gain over a random partner against the
same gain on null networks.}
\label{tab:bilateral}
\setlength{\tabcolsep}{4pt}
\begin{tabular}{llcccccc}
\toprule
PDE & concept & single & +opposite & +same sign & +random & gain & null $\pnf$ \\
\midrule
Burgers & $u$ & 0.585 & 0.786 & 0.679 & 0.629 & 0.157 & 0.272 \\
Allen--Cahn & $u$ & 0.410 & 0.583 & 0.450 & 0.442 & 0.141 & 0.204 \\
Convection $\beta=1$ & $u$ & 0.416 & 0.761 & 0.484 & 0.460 & 0.301 & 0.215 \\
 & $x-t$ & 0.541 & 0.863 & 0.591 & 0.585 & 0.278 & 0.282 \\
Convection $\beta=30$ & $u$ & 0.041 & 0.081 & 0.049 & 0.047 & 0.034 & 0.005 \\
Schr\"odinger & $\mathrm{Re}\,\psi$ & 0.354 & 0.785 & 0.397 & 0.390 & 0.395 & 0.345 \\
 & $\mathrm{Im}\,\psi$ & 0.591 & 0.753 & 0.641 & 0.611 & 0.142 & 0.125 \\
\bottomrule
\end{tabular}
\end{table}

\subsection{SAE design: dictionary size, sparsity and variants}
\label{sec:ablations}

\paragraph{Atoms and probes.}
A single atom usually captures part of what the full layer holds. For the Burgers shock gradient the probe reaches
$R^2=0.85$ while the best atom has $|r|=0.41$; for the KdV gradient the gap between probe $R^2$ and best-atom $r^2$ is
0.52, as predicted in advance. The Fisher--KPP front is the exception, with a single atom at $|r|=0.98$ for $|u_x|$. Atoms
are best read as readable views onto a distributed representation, and for some features one atom is the whole story.

\paragraph{Dictionary size and sparsity.}
Tables~\ref{tab:overcompleteness} and~\ref{tab:sparsity} vary $D$ and $\lambda$. Alignment changes little with dictionary
size: it is best at $D=256$ for heat and Burgers and at $D=512$ for Schr\"odinger, and reconstruction error stays at
$4$--$7\times10^{-4}$ throughout. Across the sparsity sweep (mean active atoms from 253 to 29) alignment is flat at
0.57--0.58 up to the default and rises to 0.66 at $\lambda=0.1$, at the cost of a doubled reconstruction error. A sparser
SAE would therefore give somewhat cleaner atoms; we keep the default so that all other results use one setting.

\begin{table}[t]
\small\centering
\caption{Dictionary size: mean best-atom $|r|$ over the concept panel and reconstruction MSE on held-out points (PINN seed
0, SAE seed 0, $\lambda=0.02$).}
\label{tab:overcompleteness}
\setlength{\tabcolsep}{5pt}
\begin{tabular}{llccc}
\toprule
PDE & Metric & $D{=}128$ & $D{=}256$ & $D{=}512$ \\
\midrule
\multirow{2}{*}{Heat}        & mean $|r|$ & 0.737 & 0.783 & 0.744 \\
                             & MSE        & 0.0005 & 0.0004 & 0.0005 \\
\addlinespace
\multirow{2}{*}{Burgers}     & mean $|r|$ & 0.529 & 0.612 & 0.574 \\
                             & MSE        & 0.0006 & 0.0004 & 0.0005 \\
\addlinespace
\multirow{2}{*}{Schr\"odinger} & mean $|r|$ & 0.709 & 0.708 & 0.729 \\
                             & MSE        & 0.0007 & 0.0005 & 0.0006 \\
\bottomrule
\end{tabular}
\end{table}

\begin{table}[t]
\small\centering
\caption{Sparsity: mean active atoms $L_0$, reconstruction MSE and mean best-atom $|r|$ on Burgers (PINN seed 0, SAE seed
0, $D=512$). Default setting in bold.}
\label{tab:sparsity}
\setlength{\tabcolsep}{5pt}
\begin{tabular}{lccc}
\toprule
$\lambda$ & $L_0$ & MSE & mean $|r|$ \\
\midrule
$10^{-4}$ & 253 & 0.0005 & 0.578 \\
$10^{-3}$ & 250 & 0.0005 & 0.577 \\
$10^{-2}$ & 214 & 0.0004 & 0.580 \\
$\mathbf{0.02}$ & \textbf{177} & \textbf{0.0005} & \textbf{0.574} \\
$0.1$     & 68 & 0.0010 & 0.660 \\
$0.3$     & 29 & 0.0023 & 0.631 \\
\bottomrule
\end{tabular}
\end{table}

\paragraph{SAE variants.}
We scored five SAEs on the six well-trained PDEs of the development round (heat, Burgers, convection $\beta=1$,
Schr\"odinger, KdV, Fisher--KPP) with a rule fixed before the run: the number of concepts for which the best atom beats
the null of the same SAE type (Table~\ref{tab:variants}). The default L1 SAE with $D=512$ tied for the best score with a
TopK SAE on a 64-atom dictionary and was the most stable across seeds. Derivative-augmented SAEs helped on heat but not
overall, so the plain hidden state is a sufficient input.

\begin{table}[t]
\centering\small
\caption{SAE variants. Score: concepts (of 36) for which the best atom, mean of 3 PINN $\times$ 2 SAE seeds, beats the null
$\pnf$ of the same SAE type. Stability: mean correlation between top-atom activation maps of two SAE seeds. FVU: fraction
of variance unexplained.}
\label{tab:variants}
\begin{tabular}{lcccc}
\toprule
SAE & Input & FVU & Score & Stability \\
\midrule
L1, $D=512$ (default) & $\tilde h$ & 0.0004 & \textbf{23/36} & \textbf{0.781} \\
L1, $D=64$ & $\tilde h$ & 0.0009 & 21/36 & 0.727 \\
TopK $k=8$, $D=64$ & $\tilde h$ & 0.0016 & \textbf{23/36} & 0.720 \\
L1, $D=512$ & jet & 0.0004 & 21/36 & 0.693 \\
TopK $k=8$, $D=64$ & jet & 0.0026 & 16/36 & 0.771 \\
\bottomrule
\end{tabular}
\end{table}

\subsection{Feature consistency}
\label{sec:consistency}

What reproduces across SAE seeds is the behaviour of the features. Each atom has a counterpart in every other seed whose activation map
correlates with it at 0.81--0.86, and the best-aligned atoms of different seeds have correlated activation maps, from
0.48 for the failed convection PINN to 0.81 for heat (0.78 on the six well-trained PDEs of Table~\ref{tab:variants};
Table~\ref{tab:consistency}). The decoder directions themselves do not reproduce: their Hungarian-matched cosine is
0.265--0.270, close to the 0.255 of two random dictionaries. This is expected for a 512-atom dictionary on a 128-unit layer whose activity occupies only
a few effective dimensions (\S\ref{sec:collapse}), since many different sets of directions then give the same codes.
We therefore interpret atoms through their activation maps.

\begin{table}[t]
\small\centering
\caption{Consistency across 3 SAE seeds per PINN (L1 SAE, layer 5; mean over 3 PINN seeds). Decoder cosine:
Hungarian-matched cosine between decoder columns, with the value for two random dictionaries. Map correlation:
correlation between activation maps of the best-aligned atoms in different seeds, and between each atom and its best
match in another seed.}
\label{tab:consistency}
\setlength{\tabcolsep}{5pt}
\begin{tabular}{lcccc}
\toprule
PDE & Decoder cos. & Random dict. & Map corr.\ (top atoms) & Map corr.\ (best match) \\
\midrule
Heat           & 0.270 & 0.255 & 0.81 & 0.86 \\
Burgers        & 0.268 & 0.255 & 0.52 & 0.82 \\
Allen--Cahn    & 0.268 & 0.255 & 0.70 & 0.82 \\
Convection $\beta=1$  & 0.270 & 0.255 & 0.64 & 0.85 \\
Convection $\beta=30$ & 0.265 & 0.255 & 0.48 & 0.84 \\
Schr\"odinger  & 0.269 & 0.255 & 0.76 & 0.81 \\
\bottomrule
\end{tabular}
\end{table}

\subsection{What failed PINNs have and have not learned}
\label{sec:collapse}

\begin{figure}[t]
\centering
\includegraphics[width=\linewidth]{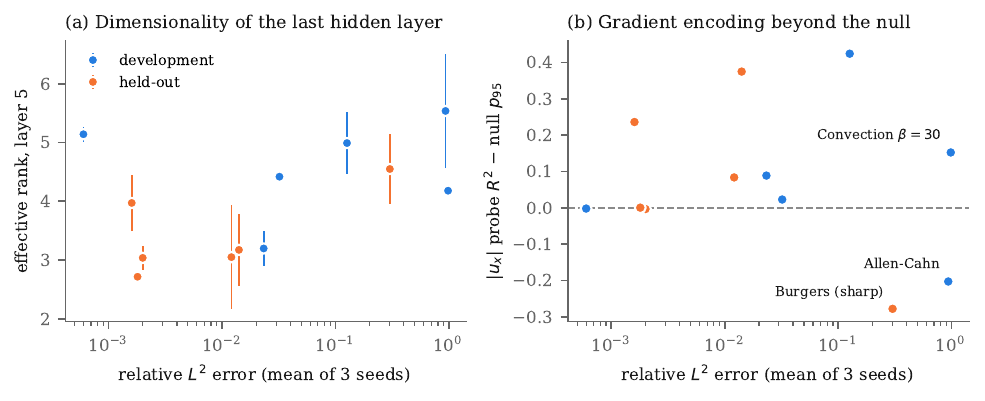}
\caption{All 36 PINNs (12 PDEs, mean $\pm$ s.d.\ over 3 seeds). (a) Effective rank of the z-scored last hidden layer.
(b) Probe $R^2$ for the gradient concept minus the null 95th percentile. Allen--Cahn and sharp Burgers fall below zero;
convection with $\beta=30$ stays above only because its null is near zero; smooth, well-trained PDEs sit at zero because
every network decodes their gradient.}
\label{fig:collapse}
\end{figure}

\physsae{} also shows what a PINN that fails to solve its equation has learned. Failed PINNs do not encode the
derivative structure of their solution beyond the null (Figure~\ref{fig:collapse}b): the gradient concept is decoded
worse than by null networks in the Allen--Cahn and sharp Burgers PINNs (for the latter, 0.537 against 0.815), and with
a probe the Allen--Cahn PINN beats the null on no concept. The failed convection PINN with $\beta=30$ keeps only a faint
trace of its solution (probe $R^2$ 0.16--0.20 for amplitude, gradient and signed $u$, against null $\pnf\le0.042$). Its
best atom for the characteristic coordinate $x-30t$ has $|r|=0.96$, but on $[-1,1]\times[0,1]$ this coordinate is
dominated by $-30t$, and its partial correlation after removing $t$ is $-0.04$, so the atom encodes time. The
diagnostic value lies in this contrast: for a well-trained PINN the SAE recovers the solution's derivatives and
operator terms; for a failed one it recovers coordinates.

The dimensionality of the last hidden layer does not separate the two cases. Its effective rank (the exponential of the
entropy of the normalised covariance eigenvalues~\citep{roy2007effective}) lies between 2.7 and 5.5 for all 12 PDEs, well below
the 128 units, with no consistent relation to error (Figure~\ref{fig:collapse}a).

\section{Discussion}
\label{sec:discussion}

\paragraph{What the results show.}
Sparse autoencoders turn the hidden layers of a PINN into a set of readable features, and the null comparison tells
us which of those features training produced. The clearest result is that trained PINNs represent the individual terms
of their own operator, which networks trained on other equations do not, and that this held for three equations chosen
after the analysis was fixed. The second-derivative terms are the strongest evidence, since they cannot be reconstructed
from the PINN's output alone. SAE atoms expose part of this structure as individual features that can be plotted,
compared and removed, while a linear probe on the full layer recovers more of it. Removing an aligned atom changes the
solution mainly where its concept is large and changes the residual differently from comparable atoms.

\paragraph{Terminology.}
We use \emph{alignment} for correlation between an atom's activation map and a concept, \emph{beats the null} for
alignment beyond what untrained and differently trained networks show, and \emph{causal evidence} for changes in
output or residual after editing the original hidden state. We do not claim that atoms \emph{are} physical variables;
we claim that they are aligned with physical concepts beyond chance and have a measurable role in the PINN's computation.

\paragraph{Limitations.}
Several results set the scope of these claims. (1)~The footprints of SAE atoms are concentrated, but at layer 4
output-matched atoms from the same SAE are about as concentrated as concept-aligned ones (0.140 against 0.151), and the
activation-matched control gap $\Delta_{\mathrm{ctrl}}$ at layer 5 is positive only for heat and Schr\"odinger. Concentration is therefore a
property of sparse codes, and concept specificity has to be read from support capture and residual effects. (2)~The
direction of the residual effect depends on the equation and the layer. A prediction that concept-aligned atoms would damage the
residual more than controls held weakly for KdV and was reversed for Fisher--KPP, and asking which operator term changes
after a term-aligned atom is removed did not settle the sign: the aligned term changed most in about half of the cases,
no more often than for controls, as expected when every term depends on $u$. (3)~Decoder directions are not
reproducible across SAE seeds, only activation maps are. (4)~For smooth solutions every network decodes most concepts at
$R^2\approx1$, so such PDEs cannot test representational claims. (5)~Not every prediction fixed in advance held: besides the residual
sign for Fisher--KPP, the KdV gradient probe (0.994) did not exceed its null (0.997) at the smooth-solution ceiling, and the
Fisher--KPP gradient was carried almost entirely by one atom rather than spread across many. The predictions were also
written by the authors of the development analysis, and an independent replication would be more convincing. (6)~All PDEs are one-dimensional, all PINNs share one $5\times128$ $\tanh$
architecture and one training method, probes are linear, and the null set of 20--23 networks mixes untrained and
differently trained ones. Extending the analysis to 2D and 3D problems and to larger networks, and automating concept
discovery (for example with symbolic regression on atom activations), are natural next steps.

\paragraph{Recommendations.}
For SAE analyses of scientific networks we recommend comparing each statistic with the same statistic on untrained and
differently trained networks, measuring intervention footprints at a layer where they are not proportional to the
activation, comparing similar-looking atoms through selectivity and concept overlap, and pre-registering the claims the
conclusions rest on.

\section{Conclusion}
\label{sec:conclusion}

\physsae{} brings sparse autoencoders and direct hidden-state interventions to physics-informed neural networks. On 12
PDEs it finds atoms that follow shocks, interfaces, solitons and fronts, shows with null networks and a pre-registered
test that trained PINNs encode the terms of the PDE they solve, and shows that removing aligned atoms changes the
solution where their concepts live. Failed PINNs lose this structure and fall back to coordinates. Together these
results make the internal representations of PINNs readable and give a concrete way to check what a PINN has learned.

\subsubsection*{Acknowledgements}
We thank Rafa\l{} Wrona for reviewing the manuscript. We are also grateful to Prof.\ Shreyus, Prof.\ Rajesh, Mayank, Omkar and Tabassum for their support as
members of the QuaNad Research Laboratory at PES University.

\bibliography{references}
\bibliographystyle{iclr2027_conference}

\clearpage
\appendix

\section{Additional atom galleries}
\label{app:atoms}

\begin{figure}[!htbp]
\centering
\begin{subfigure}[t]{0.49\linewidth}\centering
\includegraphics[width=\linewidth]{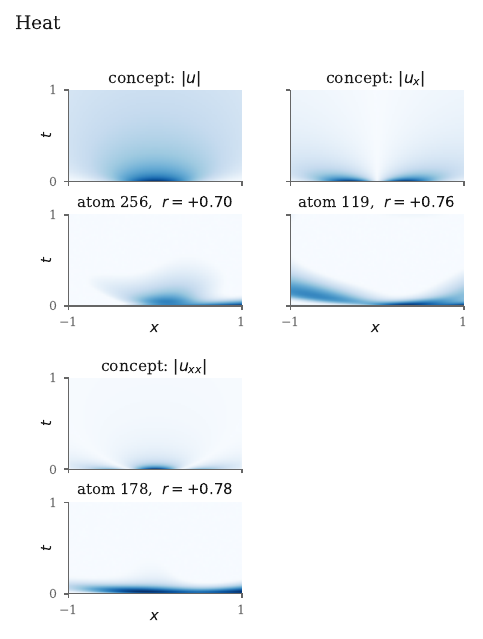}\caption{Heat}\end{subfigure}\hfill
\begin{subfigure}[t]{0.49\linewidth}\centering
\includegraphics[width=\linewidth]{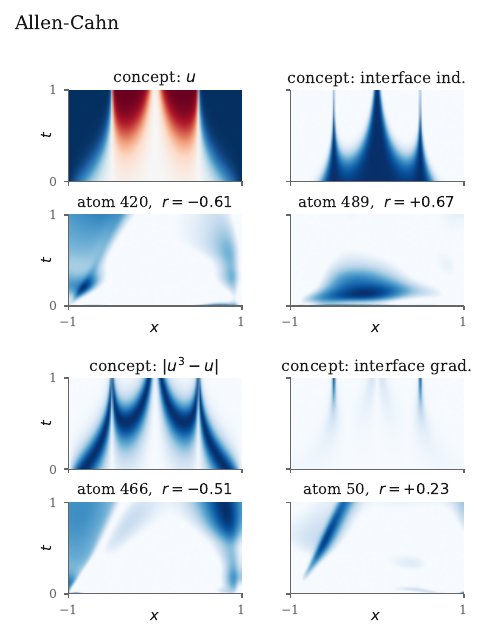}\caption{Allen--Cahn (failed PINN)}\end{subfigure}
\caption{Atom galleries for heat and Allen--Cahn, as in Figure~\ref{fig:atoms}.}
\end{figure}

\begin{figure}[!htbp]
\centering
\begin{subfigure}[t]{0.49\linewidth}\centering
\includegraphics[width=\linewidth]{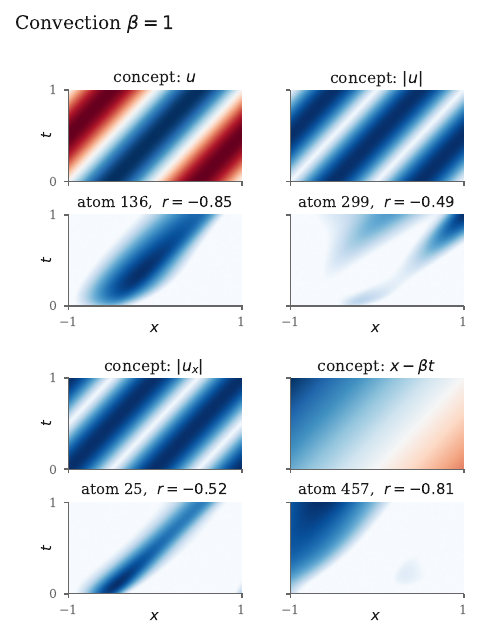}\caption{Convection, $\beta=1$}\end{subfigure}\hfill
\begin{subfigure}[t]{0.49\linewidth}\centering
\includegraphics[width=\linewidth]{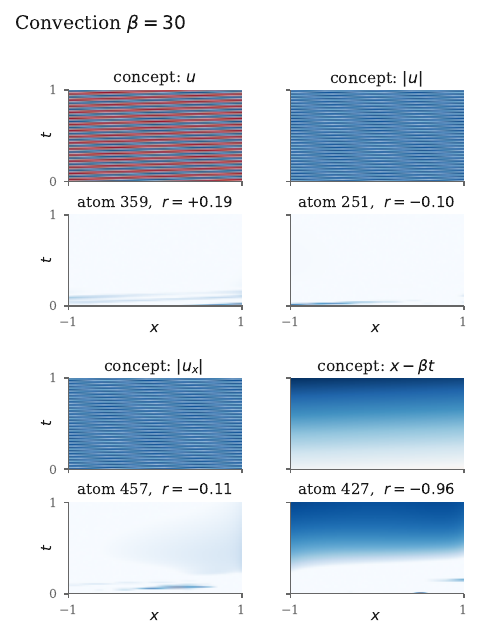}\caption{Convection, $\beta=30$ (failed PINN)}\end{subfigure}
\caption{Atom galleries for convection. For $\beta=30$ the best atom follows time rather than the solution
(\S\ref{sec:collapse}).}
\end{figure}

\begin{figure}[!htbp]
\centering
\begin{subfigure}[t]{0.49\linewidth}\centering
\includegraphics[width=\linewidth]{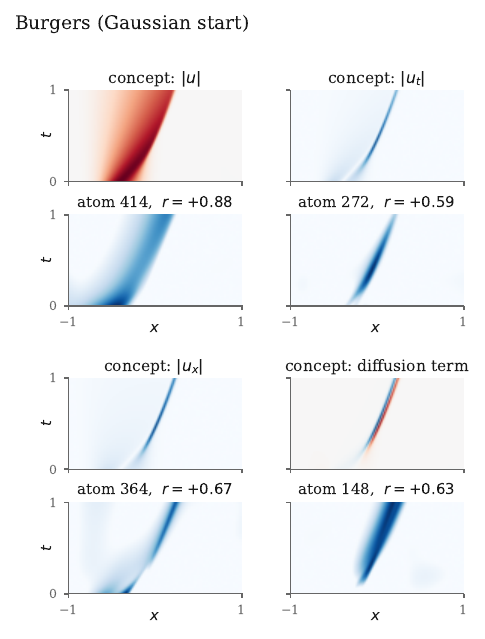}\end{subfigure}
\caption{Atom gallery for Burgers with a Gaussian initial condition (held-out), including the atom assigned to the
diffusion term, which is localised on the forming shock.}
\end{figure}

\clearpage
\section{Additional tables}
\label{app:tables}

\begin{table}[!htbp]
\centering\small
\caption{Development PDEs: trained PINN against null $\pnf$. SAE: best-atom $|r|$, mean of 3 PINN $\times$ 3 SAE seeds.
Probe: ridge $R^2$ at layer 5, mean of 3 PINN seeds. $>$ marks values above the null (ties at 1.000 resolved on
unrounded values).}
\label{tab:app_null}
\begin{tabular}{llcccccc}
\toprule
PDE & concept & SAE & $\pnf$ & & probe & $\pnf$ & \\
\midrule
Heat & $|u|$ & 0.788 & 0.841 & & 1.000 & 1.000 & $>$ \\
 & $|u_x|$ & 0.751 & 0.843 & & 0.996 & 0.998 & \\
 & $|u_{xx}|$ & 0.800 & 0.861 & & 0.991 & 0.991 & \\
Burgers & $|u|$ & 0.627 & 0.751 & & 0.997 & 0.990 & $>$ \\
 & $|u_x|$ & 0.411 & 0.401 & $>$ & 0.850 & 0.425 & $>$ \\
 & $|u_{xx}|$ & 0.512 & 0.483 & $>$ & 0.862 & 0.620 & $>$ \\
 & $|u_t|$ & 0.536 & 0.559 & & 0.903 & 0.847 & $>$ \\
 & $u$ & 0.800 & 0.800 & & 0.998 & 0.988 & $>$ \\
Allen--Cahn & $|u|$ & 0.687 & 0.746 & & 0.908 & 0.963 & \\
 & interface ind. & 0.692 & 0.747 & & 0.831 & 0.932 & \\
 & interface grad. & 0.270 & 0.333 & & 0.323 & 0.526 & \\
 & $|u^3-u|$ & 0.458 & 0.449 & $>$ & 0.773 & 0.887 & \\
 & $u$ & 0.643 & 0.707 & & 0.947 & 0.976 & \\
Conv.\ $\beta=1$ & $|u|$ & 0.529 & 0.433 & $>$ & 0.995 & 0.972 & $>$ \\
 & $|u_x|$ & 0.515 & 0.395 & $>$ & 0.993 & 0.970 & $>$ \\
 & $u$ & 0.807 & 0.654 & $>$ & 1.000 & 1.000 & $>$ \\
 & $x-t$ & 0.825 & 0.883 & & 1.000 & 1.000 & \\
Conv.\ $\beta=30$ & $|u|$ & 0.097 & 0.055 & $>$ & 0.160 & 0.010 & $>$ \\
 & $|u_x|$ & 0.107 & 0.055 & $>$ & 0.161 & 0.009 & $>$ \\
 & $u$ & 0.205 & 0.077 & $>$ & 0.198 & 0.042 & $>$ \\
 & $x-30t$ & 0.941 & 0.863 & $>$ & 0.999 & 1.000 & \\
Schr\"odinger & $|\psi|$ & 0.848 & 0.790 & $>$ & 1.000 & 0.994 & $>$ \\
 & $\mathrm{Re}\,\psi$ & 0.714 & 0.614 & $>$ & 1.000 & 0.973 & $>$ \\
 & $\mathrm{Im}\,\psi$ & 0.769 & 0.722 & $>$ & 1.000 & 0.975 & $>$ \\
 & $|\partial_x|\psi||$ & 0.733 & 0.762 & & 0.997 & 0.908 & $>$ \\
\midrule
\multicolumn{2}{l}{Above null} & \multicolumn{3}{c}{13/25} & \multicolumn{3}{c}{16/25} \\
\bottomrule
\end{tabular}
\end{table}

\begin{table}[!htbp]
\centering\small
\caption{Atom distinctness (layer 5, SAE seed 0, distinct atoms assigned per concept; mean over concepts and 3 PINN
seeds). Selectivity: $|r|$ with the own concept minus the largest $|r|$ with another concept in the panel. Partial: $|r|$
with the own concept after regressing out the other concepts. Own best: cases where the atom matches its own concept at
least as well as any other.}
\label{tab:app_distinct}
\setlength{\tabcolsep}{4pt}
\begin{tabular}{lcccc}
\toprule
PDE & mean $|r|$ (own) & selectivity & partial & own best \\
\midrule
Heat           & 0.778 & 0.143 & 0.694 & 9/9 \\
Burgers        & 0.569 & 0.237 & 0.557 & 9/12 \\
Allen--Cahn    & 0.518 & 0.112 & 0.552 & 7/12 \\
Convection $\beta=1$  & 0.651 & 0.184 & 0.571 & 9/12 \\
Convection $\beta=30$ & 0.332 & 0.133 & 0.309 & 3/12 \\
Schr\"odinger  & 0.771 & 0.069 & 0.701 & 9/12 \\
Burgers (Gaussian) & 0.692 & 0.158 & 0.433 & 9/12 \\
\bottomrule
\end{tabular}
\end{table}

\begin{table}[!htbp]
\centering\small
\caption{Operator terms in the development PDEs, KdV and Fisher--KPP: probe $R^2$ at layers 3 and 5 against null $\pnf$, and best-atom
$|r|$ against its null. Several passes for smooth solutions are ties near the ceiling.}
\label{tab:app_terms_dev}
\begin{tabular}{llccc}
\toprule
PDE & term & probe h3 / h5 & null $\pnf$ & SAE $|r|$ / $\pnf$ \\
\midrule
Burgers & $uu_x$ & 0.057 / 0.450 & 0.074 & 0.423 / 0.394 \\
 & $\nu u_{xx}$ & 0.027 / 0.657 & 0.072 & 0.506 / 0.470 \\
Heat & $u_{xx}$ & 0.877 / 1.000 & 0.998 & 0.801 / 0.874 \\
Conv.\ $\beta=30$ (failed) & $\beta u_x$ & 0.123 / 0.197 & 0.043 & -- \\
Schr\"odinger & dispersion & 0.999 / 1.000 & 0.989 & 0.847 / 0.773 \\
 & $|\psi|^2\psi$ & 0.998 / 1.000 & 0.930 & 0.832 / 0.651 \\
KdV & $6uu_x$ & 0.946 / 1.000 & 1.000 & 0.809 / 0.723 \\
 & $u_{xxx}$ & 0.875 / 1.000 & 0.998 & 0.833 / 0.710 \\
Fisher--KPP & $Du_{xx}$ & 0.681 / 1.000 & 0.993 & 0.775 / 0.809 \\
 & $ru(1-u)$ & 0.944 / 1.000 & 0.999 & 0.984 / 0.885 \\
\bottomrule
\end{tabular}
\end{table}

\begin{table}[!htbp]
\centering\small
\caption{Held-out Burgers, modified Burgers and advection--diffusion, all concepts: probe $R^2$ at layers 3/4/5, null $\pnf$ at layer 5, probe from cubic polynomial features of
$(x,t,\hat u)$ (exploratory), and best SAE atom $|r|$ against its null.}
\label{tab:app_phase6}
\begin{tabular}{llcccc}
\toprule
PDE & concept & probe h3/h4/h5 & null $\pnf$ & poly$(x,t,\hat u)$ & SAE / $\pnf$ \\
\midrule
Burgers (Gauss.) & $|u|$ & 0.974/0.999/1.000 & 0.988 & 1.000 & 0.890/0.837 \\
 & $|u_x|$ & 0.432/0.809/0.994 & 0.619 & 0.858 & 0.644/0.438 \\
 & $|u_{xx}|$ & 0.465/0.800/0.915 & 0.680 & 0.517 & 0.682/0.520 \\
 & $|u_t|$ & 0.449/0.785/0.989 & 0.586 & 0.827 & 0.624/0.472 \\
 & $uu_x$ & 0.387/0.792/0.993 & 0.558 & 0.836 & 0.610/0.465 \\
 & $\nu u_{xx}$ & 0.077/0.747/0.991 & 0.149 & 0.361 & 0.682/0.520 \\
Mod.\ Burgers & $|u|$ & 0.982/0.999/1.000 & 0.998 & 1.000 & 0.914/0.886 \\
 & $|u_x|$ & 0.582/0.914/0.997 & 0.914 & 0.913 & 0.662/0.619 \\
 & $|u_{xx}|$ & 0.569/0.849/0.925 & 0.808 & 0.656 & 0.747/0.669 \\
 & $|u_t|$ & 0.639/0.890/0.994 & 0.860 & 0.851 & 0.705/0.694 \\
 & $u^2u_x$ & 0.414/0.855/0.996 & 0.819 & 0.812 & 0.692/0.660 \\
 & $\nu u_{xx}$ & 0.194/0.711/0.988 & 0.616 & 0.584 & 0.747/0.669 \\
Adv.--diff. & $|u|$ & 0.998/1.000/1.000 & 0.996 & 1.000 & 0.941/0.936 \\
 & $|u_x|$ & 0.772/0.977/0.999 & 0.763 & 0.963 & 0.863/0.570 \\
 & $|u_{xx}|$ & 0.596/0.871/0.979 & 0.651 & 0.490 & 0.716/0.532 \\
 & $|u_t|$ & 0.759/0.968/0.999 & 0.747 & 0.964 & 0.856/0.562 \\
 & $au_x$ & 0.772/0.977/0.999 & 0.763 & 0.963 & 0.863/0.570 \\
 & $Du_{xx}$ & 0.282/0.801/0.998 & 0.224 & 0.703 & 0.716/0.532 \\
\bottomrule
\end{tabular}
\end{table}

\begin{table}[!htbp]
\centering\small
\caption{Propagated interventions by layer (L1 SAE; mean over concepts and 3 PINN seeds). $\ESF$ and support capture
(SC) for the top atom, its output-matched controls and the best-aligned PCA component.}
\label{tab:app_layers}
\setlength{\tabcolsep}{4pt}
\begin{tabular}{lcccccc}
\toprule
Layer & $\ESF$ top & $\ESF$ ctrl & $\ESF$ PCA & SC top & SC ctrl & SC PCA \\
\midrule
3 & 0.143 & 0.135 & 0.333 & 0.481 & 0.239 & 0.319 \\
4 & 0.151 & 0.140 & 0.387 & 0.486 & 0.256 & 0.315 \\
5 & 0.177 & 0.170 & 0.502 & 0.480 & 0.213 & 0.269 \\
\bottomrule
\end{tabular}
\end{table}

\clearpage
\section{Hyperparameters and implementation}
\label{app:hyperparams}

\begin{table}[!htbp]
\centering\small
\caption{Hyperparameters.}
\begin{tabular}{p{0.42\linewidth}p{0.52\linewidth}}
\toprule
Setting & Value \\\midrule
\multicolumn{2}{l}{\textit{PINN}}\\
Hidden layers / width & 5 / 128, $\tanh$, Xavier normal \\
Adam iterations / learning rate / clipping & 8000 / $10^{-3}$ / norm 1.0 \\
L-BFGS & $\leq300$ iterations, strong Wolfe; reverted if the loss increases \\
Loss weights $w_{\mathrm{BC}}, w_{\mathrm{IC}}, w_F$ & 100, 100, 1 \\
Boundary / initial / residual points per step & 150 / 150 / 3000 \\
PINN seeds & 0, 1, 2 \\\midrule
\multicolumn{2}{l}{\textit{Sparse autoencoder}}\\
Hidden width $d_h$ / dictionary size $D$ & 128 / 512 ($4\times$) \\
Sparsity $\lambda$ & $2\times10^{-2}$ \\
Epochs / batch / learning rate & 600 / 8192 / $10^{-3}$ \\
SAE seeds per PINN & 3 (2 for the variant comparison) \\\midrule
\multicolumn{2}{l}{\textit{Evaluation}}\\
Grid $n_x \times n_t$ & $200 \times 100$; 80/20 random split \\
Probe & ridge, penalty by GCV over $10^{-3}$--$10^{3}$ \\
Null networks & 5 untrained + 15 (development) or 18 (held-out) PINNs of other PDEs; 95th percentile \\
Matched controls & 10 per top atom \\
Interventions & $\alpha=1$; layers 3, 4, 5 \\
Bilateral random partners & 20 \\
\bottomrule
\end{tabular}
\end{table}

\paragraph{Reference solutions.} The NLS soliton has amplitude 2 and speed 2 on $[-5,5]$; the KdV soliton has speed
$c=2$ on $[-5,5]$; Fisher--KPP uses $D=0.01$, $r=5$; advection--diffusion uses $a=0.5$, $D=0.002$. The held-out Burgers
references stay within $3\times10^{-3}$ of zero at $x=\pm1$, consistent with the homogeneous Dirichlet condition used in
training. All figures and tables were generated from the stored result files by a single script released with the
code.

\end{document}